%% file: main.tex
\documentclass[10pt,twocolumn,letterpaper]{article}

\usepackage[pagenumbers]{wacv} % To force page numbers, e.g. for an arXiv version

\input{preamble}

\definecolor{wacvblue}{rgb}{0.21,0.49,0.74}
\usepackage[pagebackref,breaklinks,colorlinks,allcolors=wacvblue]{hyperref}

\def\wacvPaperID{1085}
\def\confName{WACV}
\def\confYear{2027}

\title{Exploiting Biases in Attention for Transformers}

\author{
\begin{tabular}{c}
Charles Herr$^{1}$ \quad Gabriele Spadaro$^{2}$ \quad Enzo Tartaglione$^{1,3}$
\end{tabular}
\\~\\
$^1$LTCI, T\'el\'ecom Paris, Institut Polytechnique de Paris, France~~ \\ $^2$University of Turin, Italy \qquad $^3$CNR-ISASI, Italy~~ \\
 {\tt\small enzo.tartaglione@telecom-paris.fr}
}

\begin{document}
\maketitle
\input{sec/0_abstract}    
\input{sec/1_intro}
\input{sec/2_related_work}
\input{sec/3_method}
\input{sec/4_experiments}
\input{sec/5_conclusion}

\section*{Acknowledgments} This work benefited from Hi! PARIS and State funding managed by the French National Research Agency (ANR) under the France 2030 program, reference ANR-23-IACL-0005. This work was also supported by the Italian MUR through
the FIS 2 project FIS-2023-03417 (DD n. 23314 10-12-2024, CUP D53C25002040001) and by the DARE SGA1 project, which has received funding from the European High-Performance Computing Joint Undertaking (JU) under grant agreement No. 101202459.
{
    \small
    \bibliographystyle{ieeenat_fullname}
    \bibliography{main}
}

\clearpage

\appendix

\input{sec/appendix}

\end{document}

%% file: preamble.tex
\usepackage{graphicx}
\DeclareMathOperator{\softmax}{softmax}

\usepackage{tabularx}
\usepackage{array} % Requis pour redéfinir les colonnes
\newcolumntype{C}{>{\centering\arraybackslash}X}

\usepackage{booktabs}
\usepackage{makecell}

\usepackage{xcolor}

\usepackage{pifont}
\newcommand{\cross}{\ding{55}}

\usepackage{algorithm}
\usepackage{algpseudocode}

\usepackage{tikz}
\usetikzlibrary{positioning, arrows.meta, fit, backgrounds, calc}

\usepackage{multirow} 

\DeclareMathOperator*{\TopK}{Top\text{-}k}
\newcommand{\kept}{\mathcal{K}}        % set of kept (unpruned) columns
\newcommand{\onevec}{\mathbf{1}}        % all-ones vector
\usepackage[normalem]{ulem}
\usepackage[accsupp]{axessibility}  

\newcommand\blfootnote[1]{%
  \begingroup
  \renewcommand\thefootnote{}\footnote{#1}%
  \addtocounter{footnote}{-1}%
  \endgroup
}

%% file: sec/0_abstract.tex
\begin{abstract}
Attention is known to be sparse in vision transformers. In particular, we analyze the unequal relevance of attention-matrix columns to the final computation and show that this phenomenon becomes increasingly pronounced with network depth. This can be interpreted as saying that tokens have an absolute importance for the attention process, independently of queries. Inspired by this, we propose a reformulation of the attention formula that makes this phenomenon explicit by adding an input-dependent bias to the query-key dot product. We show that this new term gives each token a scalar value indicating its anticipated influence on other tokens during the attention process, and can thus serve as a proxy for pruning attention maps. Across multiple transformer model sizes, this allows us to substantially reduce the computational cost of attention, in terms of both FLOPs and inference latency, while retaining most of the original classification accuracy. We publicly released the code and models at \href{https://github.com/EIDOSLAB/AttentionWithBias}{https://github.com/EIDOSLAB/AttentionWithBias}.
\blfootnote{This article has been accepted for publication at the 2027 IEEE/CVF Winter Conference on Applications of Computer Vision (WACV 2027).}
\end{abstract}

%% file: sec/1_intro.tex
\section{Introduction}
\label{sec:intro}

Since the introduction of the vision transformer (ViT)~\cite{dosovitskiy2021imageworth16x16words}, computer vision has been shifting away from traditional CNN-based architectures~\cite{DBLP:journals/corr/HeZRS15} to embrace the transformer architecture originally developed for natural language processing~\cite{DBLP:journals/corr/VaswaniSPUJGKP17}. Transformers now achieve strong results across a wide range of visual tasks, including image classification~\cite{dosovitskiy2021imageworth16x16words}, object detection~\cite{DBLP:journals/corr/abs-2010-04159}, semantic segmentation~\cite{DBLP:journals/corr/abs-2105-15203}, and image generation~\cite{peebles2023scalablediffusionmodelstransformers}. Their reach extends well beyond purely visual problems: by tokenizing images into sequences compatible with language models, the same architecture underpins today's large vision-language models, which jointly process visual and textual tokens for tasks such as visual question answering and instruction following~\cite{liu2023visual}. This versatility has made the transformer a near-universal backbone for modern multimodal perception.

\begin{figure}
    \centering
    \includegraphics[width=1\linewidth]{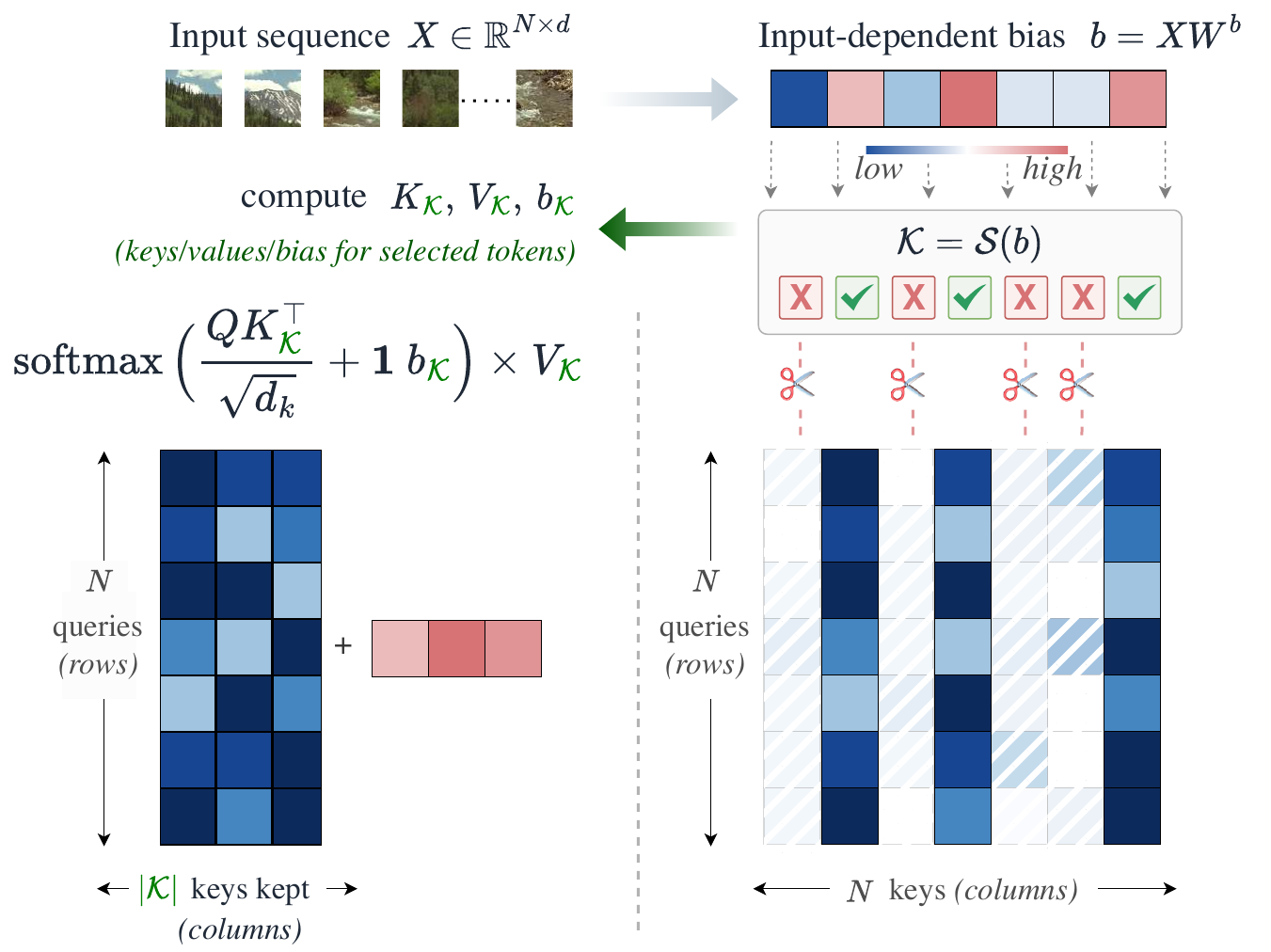}
    \caption{Bias-based column pruning. The learned bias provides a query-independent importance score for each token, allowing low-bias columns to be removed before computing keys and values. Attention is then evaluated only over the retained set \(\mathcal{K}\), reducing its size from \(N\times N\) to \(N\times|\mathcal{K}|\).}
    \label{fig:teaser}
\end{figure}

This success, however, comes at a price: the self-attention mechanism at the core of transformers has a computational cost quadratic in the length of the processed sequence. The issue is particularly acute in vision, since for square images the number of tokens grows quadratically with the image resolution, and it is amplified further in multimodal settings, where long visual token streams are concatenated with text~\cite{liu2023visual}. 
Reducing the cost of attention is thus an important and active research problem.

Existing approaches fall broadly into two families. The first rewrites the attention operator to lower its asymptotic cost, for instance by limiting attention to local windows~\cite{DBLP:journals/corr/abs-2103-14030} or replacing the softmax with a linear approximation~\cite{DBLP:journals/corr/abs-2009-14794, DBLP:journals/corr/abs-2006-16236}, usually at the expense of some accuracy. The second keeps standard attention but shortens the input sequence by suppressing tokens~\cite{DBLP:journals/corr/abs-2106-02034, bolya2023tokenmergingvitfaster, chen2023diffrate, alvar2025divprunediversitybasedvisualtoken}, exploiting visual redundancy. Both families rely, explicitly or implicitly, on a notion of \emph{token importance}, yet obtaining a reliable and cheap importance signal remains difficult and often requires auxiliary prediction modules and additional training objectives~\cite{DBLP:journals/corr/abs-2106-02034}.

In this work, we take a different route. We start from a simple empirical observation: in the attention maps of trained ViTs, some columns receive a far greater share of attention than others. This effect becomes more pronounced in the deeper layers, where attention maps degenerate into a few near-vertical lines. This suggests that certain tokens carry an \emph{absolute} importance for the attention process: an influence on other tokens that is largely \emph{independent of the query} but \emph{varies across heads}. The same phenomenon has recently been characterized from other angles, as \emph{attention sinks} in language models~\cite{xiao2024streamingllm, gu2025attentionsink} and as high-norm \emph{register} tokens in vision transformers~\cite{darcet2024registers, jiang2025testtimeregisters}; notably, \citet{gu2025attentionsink} show that attention sinks behave like an input-independent \emph{key bias} added inside the softmax. Whereas this line of work largely treats the effect as an artifact to be characterized or cleaned up, we argue that the underlying query-independent importance is a useful quantity that should be made explicit and exploited. As shown in \cref{fig:teaser}, we then add a learnable, per-head, input-dependent bias to the query-key scores before the softmax. This bias yields, for each token and each head, a scalar quantifying how much that token is expected to influence others. Since this score is computed \emph{before} keys and values are formed, it serves as a proxy for pruning attention maps, allowing us to skip computing keys and values for low-importance tokens.

Our contributions can be summarized as follows.
\begin{itemize}
    \item We identify and analyze a query-independent form of column sparsity in ViT attention maps, which becomes increasingly pronounced with depth (\cref{subsec:preliminaries} and \cref{fig:attention_sparsity}).

    \item We introduce a minimal reparameterization of self-attention that assigns each token a learned, input-dependent bias, and use it to prune low-importance key-value columns before they are computed (\cref{subsec:bias_definition,subsec:selection_pipeline}).

    \item We show that bias-based pruning reduces attention computation and inference latency across different model sizes (\cref{subsec:computational_gains,tab:pruning_results}), is complementary to sequence-reduction strategies (\cref{subsec:tome}), and achieves a better accuracy-latency trade-off than the evaluated structured-pruning baseline (\cref{subsec:structured_pruning}).

\end{itemize}

%% file: sec/2_related_work.tex
\section{Related works}

The quadratic cost of self-attention has motivated extensive research on efficient transformer architectures. We review the approaches most relevant to us and then connect our approach to the recently characterized phenomenon of \emph{attention sinks} and \emph{register tokens}.

\paragraph{Improving efficiency in vision transformers.}
The computational cost of ViTs can be reduced by targeting different sources of redundancy. Some methods modify the attention operator itself, for example by restricting interactions to local windows~\cite{DBLP:journals/corr/abs-2103-14030} or replacing softmax attention with linear approximations~\cite{DBLP:journals/corr/abs-2009-14794,DBLP:journals/corr/abs-2006-16236}, at the expense of some accuracy.
Other methods instead exploit redundancy in the token sequence. Tokens can be removed using learned or attention-based importance scores~\cite{DBLP:journals/corr/abs-2106-02034}, merged with similar tokens to preserve more information~\cite{bolya2023tokenmergingvitfaster}, or jointly pruned and merged with learned layer-wise compression rates~\cite{chen2023diffrate,mao2025pruneandmerge}. 
A further direction reduces the model itself through structured pruning, removing parameters or components~\cite{fang2023depgraph, song_pruning, lobster, structured_survey}; these approaches derive binary masks over attention heads or weight matrices, but these masks are fixed after training and shared across all inputs.

Our method targets a different granularity of redundancy. Compared with structured pruning~\cite{fang2023depgraph,song_pruning,lobster,structured_survey}, it adapts the selection to each input and attention head rather than permanently removing fixed model components. Compared with token-sequence reduction methods such as ToME and DiffRate~\cite{bolya2023tokenmergingvitfaster,chen2023diffrate}, it retains the full residual sequence and all queries. Moreover, its pruning score emerges directly from the attention reparameterization, without introducing auxiliary prediction modules specifically trained for token selection, as required by approaches such as DynamicViT~\cite{DBLP:journals/corr/abs-2106-02034}. This distinct granularity makes our method complementary to sequence-reduction techniques and, as demonstrated by its combination with ToME, improves the accuracy-efficiency trade-off in low-complexity regimes.

\paragraph{Attention sinks and register tokens.}
Our method is motivated by the observation that a small subset of attention columns consistently receives high scores across many queries, especially in deeper ViT layers. We interpret these vertical patterns as evidence of a query-independent component of token importance, which our formulation makes explicit through a learned, per-head, input-dependent bias term. This interpretation connects our work to recent studies of attention sinks and register tokens.

In language models, \citet{xiao2024streamingllm} identified \emph{attention sinks}: tokens that attract a large share of attention mass largely independently of content, and showed that merely preserving them suffices to stabilize streaming inference. In vision, \citet{darcet2024registers} showed that ViTs spontaneously repurpose a few low-information background patches as high-norm ``artifact'' tokens that store global information, and proposed adding learnable \emph{register} tokens to absorb this role and produce cleaner attention maps; \citet{jiang2025testtimeregisters} subsequently reproduced this effect at test time without retraining. Most relevant to us, \citet{gu2025attentionsink} show that attention sinks emerge from softmax normalization and behave like an input-independent \emph{key bias} added inside the softmax.

Our formulation makes a related notion of key bias explicit, while making it \emph{per-head, and input-dependent}. Rather than concentrating query-independent importance into a small set of sink or register tokens, we assign every token a learned scalar bias $b$ that exposes its expected influence across queries. 
Whereas prior work mainly characterizes or mitigates this concentration phenomenon, we model it directly and repurpose it as a pre-computable signal for pruning attention columns.

%% file: sec/3_method.tex
\section{Method} 
\label{sec:method}

Our method builds on the hypothesis that attention may contain query-independent components. We introduce the underlying hypothesis and notation in \cref{subsec:preliminaries}, define and analyze the bias in \cref{subsec:bias_definition}, and present the pruning pipeline, selection strategy, and computational gains in \cref{subsec:selection_pipeline,subsec:computational_gains}. The corresponding empirical analyses are provided in \cref{subsec:prelim_experiments}.

\subsection{Preliminaries}
\label{subsec:preliminaries}

\paragraph{Multi-head self-attention.}
\label{par:review_attention}

Let \(X \in \mathbb{R}^{N \times d}\) denote a sequence of \(N\) tokens, each represented by a \(d\)-dimensional embedding. Multi-head self-attention processes \(X\) through \(h\) parallel attention heads. For the \(i^{\text{th}}\) head, queries, keys, and values are obtained through linear projections:
\begin{equation}
    Q_i = XW_i^Q, \qquad
    K_i = XW_i^K, \qquad
    V_i = XW_i^V,
\end{equation}
where \(W_i^Q, W_i^K \in \mathbb{R}^{d \times d_k}\) and
\(W_i^V \in \mathbb{R}^{d \times d_v}\).

The output of the \(i^{\text{th}}\) head is then computed as
\begin{equation}
    \operatorname{Attention}(Q_i,K_i,V_i)
    =
    \operatorname{softmax}\left(
        \frac{Q_iK_i^\top}{\sqrt{d_k}}
    \right)V_i,
    \label{eq:attention}
\end{equation}
where the softmax is applied row-wise. The resulting attention matrix assigns, for each query token, a normalized weight to every key token, and the head output is the corresponding weighted combination of the value vectors.

Finally, the outputs of all heads are concatenated and projected back to the original embedding dimension:
\begin{equation}
    \operatorname{MultiHead}(X)
    =
    \operatorname{Concat}
    \left(
        \operatorname{head}_1,\ldots,\operatorname{head}_h
    \right)W^O,
\end{equation}
where \(W^O \in \mathbb{R}^{hd_v \times d}\) and \( \operatorname{head}_i = \operatorname{Attention}(Q_i,K_i,V_i).\)

\paragraph{Working hypothesis.}
\label{par:hypothesis}
For a given head, \(Q_iK_i^T\) computes the dot-product of each query with each key. Heads, however, may specialize and ignore irrelevant tokens: in image classification, for instance, a head specialized in detecting faces does not need tokens describing the background color. We thus hypothesize that ViT attention is \emph{column-sparse}: a small subset of tokens receives most of the attention; this subset is largely independent of the query, and the effect strengthens with depth, so that deep-layer attention maps degenerate into a few vertical lines (anchor tokens). Our method makes this query-independent importance explicit. We empirically validate this hypothesis in \cref{subsec:prelim_experiments} (\cref{fig:attention_sparsity}).

\subsection{Definition and theoretical analysis of the bias in attention}
\label{subsec:bias_definition}

\paragraph{Input-dependent attention bias.}
For each attention head, we associate every token with an additional scalar score that is computed directly from its input representation.
For the \(i^{\text{th}}\) head, we define
\begin{equation}
    b_i = XW_i^b,
    \qquad
    W_i^b \in \mathbb{R}^{d \times 1}.
\end{equation}

The projection \(W_i^b\) is layer- and head-specific, with no parameter sharing, and \(b_i\in\mathbb{R}^N\) contains one input-dependent scalar per token.
Then, we incorporate these scores into the attention logits before applying the softmax:
\begin{equation}
\begin{aligned}
    \operatorname{Attention}(Q_i,K_i,&V_i,b_i)
    = \\
    &\operatorname{softmax}\left(
        \frac{Q_iK_i^\top}{\sqrt{d_k}}
        +
        \onevec_N b_i^\top
    \right)V_i,
\end{aligned}
\label{eq:bias}
\end{equation}
where \(\onevec_N \in \mathbb{R}^{N}\) is the all-ones vector. Therefore, the scalar \((b_i)_k\) is added to the \(k^{\text{th}}\) attention column for every query. Unlike the query-key score, which depends on a specific pair of tokens, \((b_i)_k\) depends only on the representation of token \(k\) and is shared across all queries within the head.

The bias projection introduces \(d\) additional parameters and \(2Nd\) FLOPs per head. Given \(d_k \gg 1\), this overhead is negligible compared with the query, key, and value projections. Since the bias participates directly in the attention computation, \(W_i^b\) is learned jointly with the remaining model parameters, without requiring an auxiliary module or tailored loss.

\paragraph{Bias interpretation.}
To clarify the role of the proposed bias, we write the standard attention logits of the \(i^{\text{th}}\) head as
\begin{equation}
    S_i = \frac{Q_iK_i^\top}{\sqrt{d_k}}.
\end{equation}
The modified logits can then be decomposed as
\begin{equation}
    \widetilde S_i
    =
    S_i + \onevec_N b_i^\top.
    \label{eq:decomp}
\end{equation}
At the level of individual entries, we can write
\begin{equation}
    (\widetilde S_i)_{qk}
    =
    (S_i)_{qk} + (b_i)_k.
\end{equation}
The two terms play complementary roles. The interaction term \((S_i)_{qk}\) measures the compatibility between query \(q\) and token \(k\), whereas \((b_i)_k\) represents a query-independent preference for token \(k\). The latter can vary across inputs and attention heads, but remains constant across all queries of the same head.

This interpretation becomes more explicit after applying the row-wise softmax. For query \(q\), the attention weight assigned to token \(k\) is
\begin{equation}
    \widetilde a_i(q)_k
    =
    \frac{
        \exp\!\left((S_i)_{qk}\right)
        \exp\!\left((b_i)_k\right)
    }{
        \sum_j
        \exp\!\left((S_i)_{qj}\right)
        \exp\!\left((b_i)_j\right)
    }.
    \label{eq:gate}
\end{equation}
Defining
\begin{equation}
    g_{ik} = \exp\!\left((b_i)_k\right),
\end{equation}
the bias can therefore be interpreted as a positive, per-token gate that reweights query-dependent attention before normalization. Tokens with large \(g_{ik}\) are consistently promoted across queries, while tokens with small \(g_{ik}\) are suppressed.

This provides an explicit parameterization of the query-independent token importance introduced in \cref{par:hypothesis}. It also connects the bias naturally to column pruning: removing column \(k\) is equivalent to setting \(g_{ik}=0\), or, in logit space, \((b_i)_k=-\infty\). In practice, we use the finite values of \(b_i\) to determine which attention columns should be retained, as described in \cref{subsec:selection_pipeline}.

\paragraph{Properties.}
The formulation above leads to three key properties, which we denote as properties~1--3 and investigate empirically in \cref{subsec:prelim_experiments}.

    {\textit{Property 1}}. Since \(b_i\) is shared across queries, the term \(\onevec_N b_i^\top\) can capture the query-independent component of the attention pattern, including the prominent vertical structures observed in deeper layers, whereas \(S_i = Q_iK_i^\top/\sqrt{d_k}\) captures query-specific variations (\cref{fig:attention_decomposition}); we expect the relative contribution of \(b_i\) to increase with depth.

    \textit{Property 2}. The bias is not constrained a priori to match any particular statistic of the query-key scores, yet in practice it learns to approximate their pre-softmax per-column mean, increasingly so with depth (\cref{fig:bias_qk_mean_scatter}). Crucially, \(b_i\) is computed before \(S_i\), so it provides a cheap, precomputable surrogate of this per-column importance that would otherwise require materializing \(QK^{\top}\).

    \textit{Property 3}. Since \(g_{ik}=\exp((b_i)_k)\) multiplicatively modulates the contribution of token \(k\) for every query inside the softmax operation, tokens with larger bias values are expected to receive more attention mass across the rows of the attention matrix, especially in layers where the query-independent component is dominant (\cref{fig:bias_attn_corr_plot}). 

The last property is central to our method, motivating the use of \(b_i\) as a proxy for pruning attention columns.

\subsection{Bias selection strategy and pipeline}
\label{subsec:selection_pipeline}

We now use the learned bias as an input-adaptive signal for pruning. 
For clarity, we present the procedure for a single attention head and omit the head index $i$. 

\paragraph{Bias as a proxy for pruning.} 
By \cref{eq:gate}, columns with a small bias contribute little to the output, so we can prune them. The key benefit is that \(b = XW^b\) is available \emph{before} \(Q,K,V\): once the kept set \(\kept \subseteq \{1,\dots,N\}\) is chosen from \(b\), we can avoid computing the keys and values of the discarded tokens and never materialize the corresponding columns. \Cref{fig:pipeline} sketches the forward pass, while the pseudo-code is provided in the supplementary material.

\input{sec/figure_diagram}

Computational benefits depend on the selection rule. Additive masking with \(-\infty\) reproduces pruning but saves no computation, since the full \(K\), \(V\), and attention matrix are still materialized. Real savings require gathering tokens beforehand, reducing attention from \(N \times N\) to \(N \times |\kept|\). Because adaptive budgets produce irregular tensor shapes requiring padding or specialized support, we use a fixed budget with input-adaptive indices and static shapes.
\paragraph{Top-$\mathbf{k}$.}
Given a target budget \(k\), we retain the columns with the largest bias values:
\begin{equation}
    \kept = \TopK(b,k).
\end{equation}
$\TopK$
selects exactly \(k\) columns for every input and head within a layer. The retained indices remain input- and head-dependent, but their number is fixed, allowing the selected tokens to be gathered before computing \(K\) and \(V\). The resulting reduced attention operation can therefore be implemented using standard dense batched operations.

Using the same \(k\) in every layer is sub-optimal. As suggested by \textit{property 1} and validated in \cref{subsec:prelim_experiments}, the bias becomes a stronger indicator of query-independent token relevance with depth. We therefore retain more columns in early layers and prune more aggressively in deeper ones. For a network with \(L\) layers, we define a keep ratio that decreases linearly from \(\rho_{\text{high}}\) to \(\rho_{\text{low}}\):
\begin{equation}
    \rho_\ell =
    \rho_{\text{high}}
    -
    (\rho_{\text{high}}-\rho_{\text{low}})
    \frac{\ell-1}{L-1},
    \qquad
    k_\ell = \lfloor \rho_\ell N \rceil,
    \label{eq:topk_schedule}
\end{equation}
for \(\ell = 1,\dots,L\), where \(\lfloor\cdot\rceil\) denotes rounding to the nearest integer. We denote such a schedule by \(\rho_{\text{high}} \searrow \rho_{\text{low}}\); for example, \(0.9 \searrow 0.1\) keeps \(90\%\) of the columns in the first layer and progressively reduces this to \(10\%\) in the last. 

\paragraph{Training with pruning.}
Pruning can be applied at inference to a model trained with \cref{eq:bias}, but this limits how aggressively columns can be removed without degrading accuracy. We therefore apply the selection rule \(\mathcal{S}\) during training as a pruning mask in the forward pass. This allows the model to adapt to the missing columns and better preserve accuracy at larger pruning ratios. Although discarded columns receive no gradient, \(W^b\) is still optimized through the bias values of the retained columns in \cref{eq:gate}.

\subsection{Computational analysis}
\label{subsec:computational_gains}
We compare standard multi-head attention with actual column pruning, assuming \(d_k=d_v=d/h\). Summing over the \(h\) heads, the query, key, value, and output projections require \(8Nd^2\) FLOPs, while computing the attention scores and aggregating the values requires \(4N^2d\). Neglecting the lower-order cost of the softmax, standard multi-head attention therefore requires
\begin{equation}
C_{\mathrm{full}} = 8Nd^2 + 4N^2d.
\end{equation}

Let \(\vert\kept\vert\) denote the number of retained columns. All \(N\) queries are preserved, whereas keys and values are computed only for the \(k\) selected tokens. The resulting cost is
\begin{equation}
C_{\mathrm{pruned}}
=
4(N+\vert~\kept~\vert)d^2 + 4N\vert\kept\vert d,
\label{eq:cost_pruned}
\end{equation}
plus \(2Ndh\) FLOPs for computing the per-head biases. This additional cost is lower order with respect to the attention projections. Consequently, the bias projection leaves the asymptotic complexity of both the forward and backward passes unchanged, introducing almost no training overhead.

We are primarily interested in the long-sequence regime \(N \gg d\), where the attention terms dominate the projection costs. In this regime, standard attention has complexity \(C_{\mathrm{full}} = \mathcal{O}(N^2d)\), while column pruning reduces it to
\begin{equation}
C_{\mathrm{pruned}} = \mathcal{O}(N~\vert \mathcal{K} \vert~d).
\end{equation}
Our method is therefore particularly beneficial for long sequences, such as those produced by high-resolution images. Accordingly, in \cref{subsec:pruning_results,subsec:tome,subsec:structured_pruning}, we evaluate both FLOPs and inference latency at an input resolution of \(1024\times1024\).

%% file: sec/figure_diagram.tex
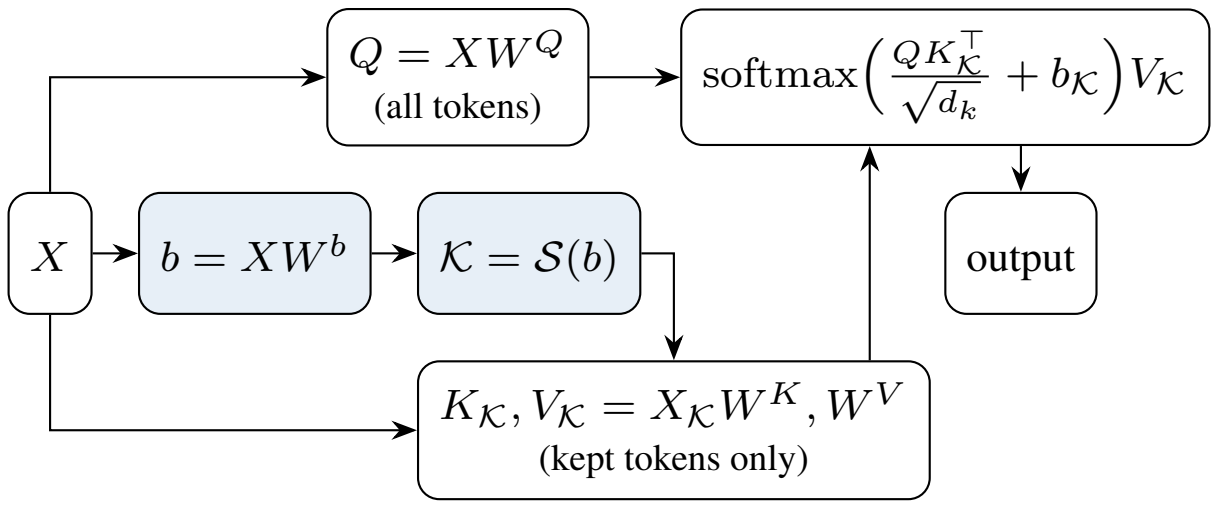
\begin{figure}[t]
    \centering
    \begin{tikzpicture}[
        >=Stealth, font=\footnotesize,
        box/.style={draw, rounded corners, align=center, inner sep=4pt, minimum height=8mm},
        hl/.style={box, fill=wacvblue!12},
        row/.style={text height=2ex, text depth=0.5ex}, % solo per i nodi della riga centrale
    ]
        \node[box, row] (x) {$X$};
        \node[hl, row, right=3mm of x] (b) {$b=XW^b$};
        \node[hl, row, right=3mm of b] (sel) {$\kept=\mathcal{S}(b)$};
        \node[box, above right=3mm and -3mm of b] (q) {$Q=XW^Q$\\\scriptsize (all tokens)};
        \node[box, below right=3mm and 3mm of b] (kv) {$K_\kept,V_\kept=X_\kept W^{K},W^{V}$\\\scriptsize (kept tokens only)};
        \node[box, right=6mm of q] (att) {$\softmax\!\Big(\tfrac{QK_\kept^\top}{\sqrt{d_k}}+b_\kept\Big)V_\kept$};
        % out: sulla riga di sel, sotto att (spostato un po' a destra)
        \node[box, row] (out) at ([xshift=5mm]att.center |- sel.center) {output};

        \draw[->] (x) -- (b);
        \draw[->] (b) -- (sel);
        \draw[->] (x.north) |- (q.west);
        \draw[->] (sel.east) -| (kv.north);
        \draw[->] (x.south) |- (kv.west);
        \draw[->] (q.east) -- (att.west);
        % kv -> att: verticale, parte dal bordo superiore di kv sotto il punto di arrivo
        \coordinate (up) at ([xshift=-5mm]att.south);
        \draw[->] (up |- kv.north) -- (up);
        % att -> out: verticale, parte dal bordo inferiore di att sopra out
        \draw[->] (out.north |- att.south) -- (out.north);
    \end{tikzpicture}
    \caption{Bias-based pruning pipeline. The importance score \(b\) is computed first (shaded) and selects the kept set \(\kept\). All queries are kept, but keys and values are computed only for the kept tokens, so the pruned columns are never materialized.}
    \label{fig:pipeline}
\end{figure}

%% file: sec/4_experiments.tex
\section{Experiments}
\label{sec:experiments}

\paragraph{Setup.}
We evaluate our method on the tiny, small, and base variants of DeiT~\cite{deit}, as well as on a ViT-L pretrained with MAE~\cite{MaskedAutoencoders2021}, all using a patch size of 16.
We append the subscript ``$b$’’ to indicate models equipped with our modified attention mechanism, resulting in DeiT-Ti$_b$, DeiT-S$_b$, DeiT-B$_b$, and ViT-L$_b$.
Our main experiments are conducted on the ImageNet ILSVRC 2012 dataset (ImageNet-1k)~\cite{5206848}, which contains approximately $1.2$ million training images and $50,000$ validation images across $1,000$ categories.\footnote{Information regarding the ImageNet license is available at \url{http://www.image-net.org/download}.}
Starting from the official pretrained checkpoints, we replace the attention module in each transformer block with the formulation introduced in Eq. \eqref{eq:bias}. For DeiT, we then fine-tune the resulting models for $70$ epochs using AdamW \cite{adamw}, a batch size of $1,024$, and an initial learning rate of $10^{-4}$ decayed according to a cosine schedule. We do not use learning-rate warmup.
For ViT-L, we start from the official MAE checkpoint fine-tuned on ImageNet-1K and fine-tune only the attention modules for $30$ epochs, using AdamW, an effective batch size of $1,024$, and, also in this case, an initial learning rate of $10^{-3}$ decayed according to a cosine schedule without learning-rate warmup.

To evaluate our strategy beyond classification, we fine-tune Mask R-CNN~\cite{he2017maskrcnn} with a DeiT-S backbone on COCO~\cite{lin2014coco} for $100$ epochs at \(1024\times1024\) resolution, following the ViTDet~\cite{li2022vitdet} training recipe. The pruned model uses the same \(0.9\searrow0.1\) schedule adopted for classification. Further details are provided in the supplementary material.

\begin{figure}[t]
\centering
\includegraphics[width=\linewidth]{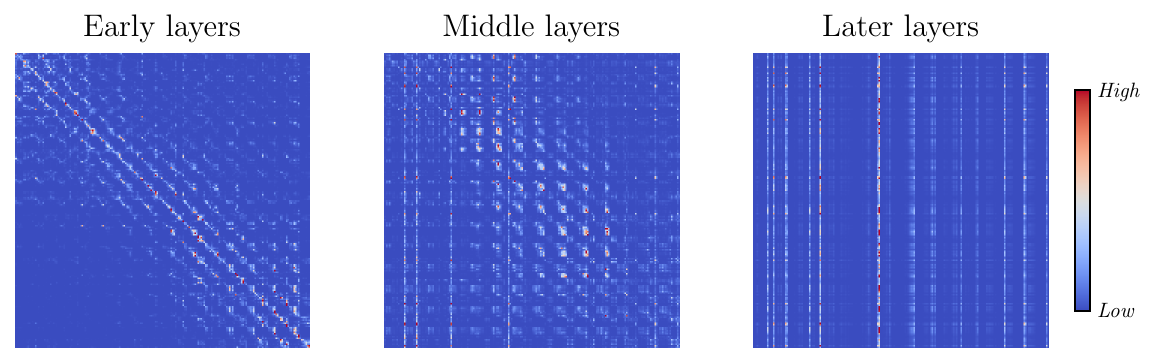}
\caption{Attention maps at different depths of the pretrained DeiT-S model. Vertical lines naturally emerge in the last layers.}
\label{fig:attention_sparsity}
\end{figure}

\subsection{Preliminary experiments}
\label{subsec:prelim_experiments}

Before evaluating pruning, we validate the assumptions underlying our method.
We analyze column sparsity using a DeiT-S pretrained on ImageNet-1K, while
experiments involving the proposed bias use a DeiT-S$_b$ fine-tuned and
evaluated with full attention, without any top-$k$ selection or pruning mask.
This isolates the behavior learned by the bias from any adaptation or implicit regularization induced by training with pruned attention columns.
In particular, neither our layer-wise pruning schedule nor an auxiliary column-ranking objective is used in these experiments.

Unless otherwise stated, visualizations refer to the same ImageNet validation image and to layers \(3\), \(7\), and \(11\) of the first attention head, representing early, middle, and late depth. The supplementary material extends these analyses to additional images, layers, and heads, as well as to the other DeiT variants and ViT-L.

\paragraph{Attention is column-sparse.}
We first qualitatively inspect the attention maps of a pretrained DeiT-S at different depths (\cref{fig:attention_sparsity}). In the early layers, attention is predominantly concentrated around the diagonal, indicating that tokens mainly interact with themselves and with spatially neighboring tokens. As depth increases, a small number of vertical structures emerge alongside the diagonal.
In the final layers, attention becomes strongly concentrated on a few columns
that consistently receive high attention across many queries. We quantify
this behavior in the supplementary material by comparing a global top-$k$
column selection with a query-dependent top-$k$ oracle.
Their gap decreases with depth and becomes nearly zero in the final
blocks.
This result supports the presence of a progressively stronger query-independent
component and motivates our depth-dependent pruning schedule.

\begin{figure}[t]
\centering
\includegraphics[width=1.0\linewidth]{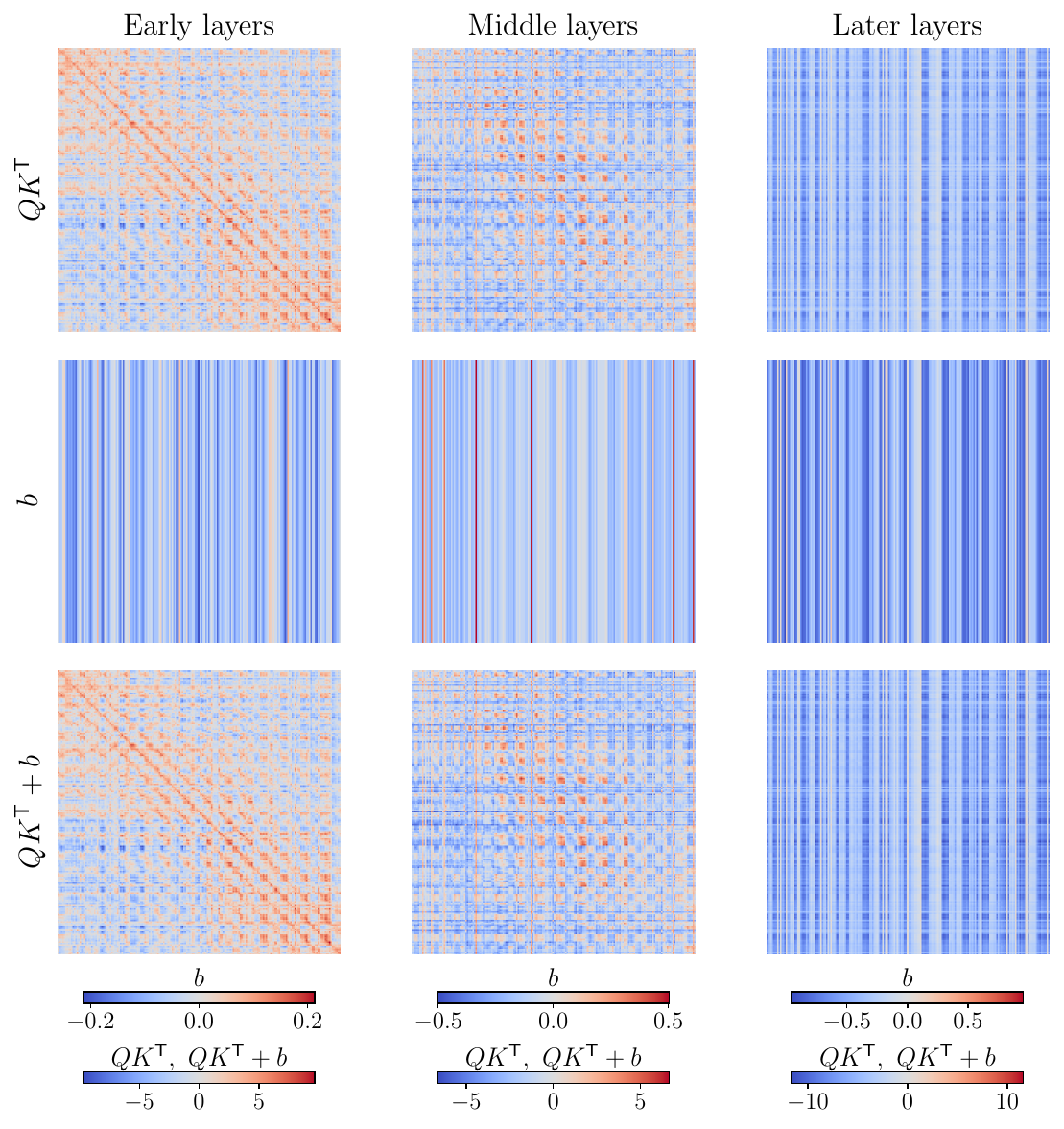}
\caption{Decomposition of the attention logits at different network depths.
}
\label{fig:attention_decomposition}
\end{figure}
\begin{figure}[t]
    \centering
    \includegraphics[width=\linewidth]{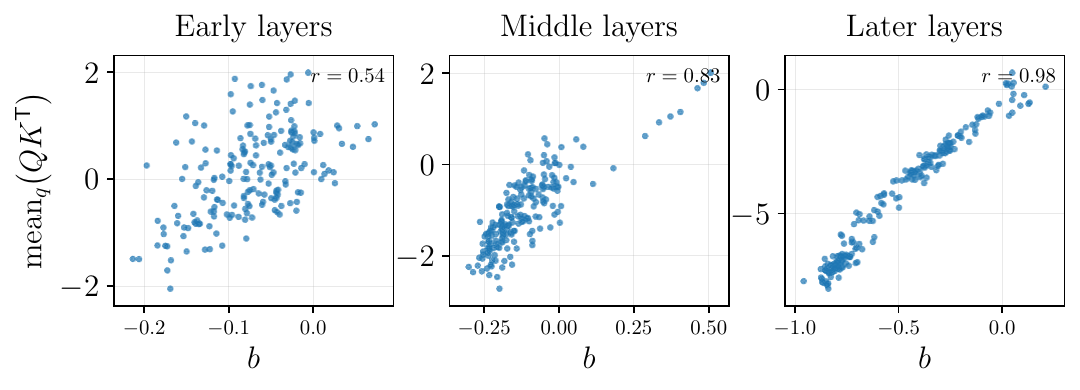}
    \caption{Learned bias \(b\) versus the per-column mean of the query-key logits \(\operatorname{mean}_q (QK^{\top})\), at three depths. 
    The correlation grows with depth (\(r\) reported per panel).
    }
    \label{fig:bias_qk_mean_scatter}
\end{figure}

\paragraph{The bias captures query-independent attention structure.}
To empirically validate
\textit{property 1} from \cref{subsec:bias_definition}, we visualize side by side the query-key logits \(QK^\top\), the bias term \(\mathbf{1}b^\top\) broadcast across queries, and their sum in \cref{fig:attention_decomposition}. The bias term captures the query-independent vertical structures, whereas the query-key term preserves query-specific patterns, including the prominent diagonal component observed in shallow layers. 

As depth increases, the vertical structures encoded by the bias become increasingly pronounced and eventually account for most of the visible structure in the combined logits, consistent with the observations in \cref{fig:attention_sparsity}. Motivated by this trend, in the supplementary material we investigate a limiting architectural variant in which the query-key term is removed from the final blocks and the attention logits are computed solely from the broadcast bias \(\mathbf{1}b^\top\).

\begin{figure}[t]
\centering
\includegraphics[width=0.9\linewidth]{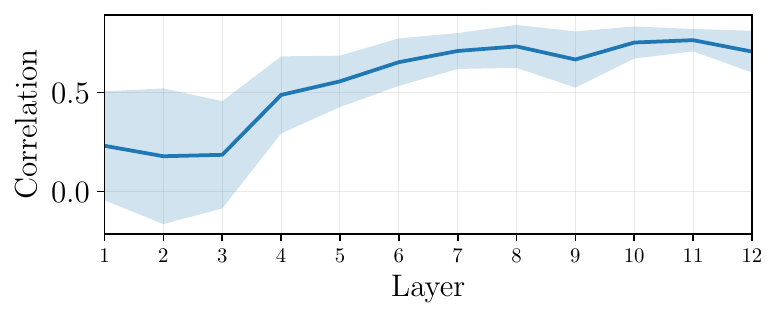}
\caption{Layer-wise Pearson correlation between the learned bias and the mean attention received by each column.} 
\label{fig:bias_attn_corr_plot}
\end{figure}

\paragraph{The bias anticipates column importance.}
We next examine 
\textit{property 2} and ask whether 
$b$ encodes information about the query-key scores. We compare $b$ with the per-column mean of $QK^\top$ across queries; the corresponding scatter plots at three depths are shown in~\cref{fig:bias_qk_mean_scatter}. A clear and increasingly tight relationship emerges with depth, suggesting that the bias learns to approximate the per-column mean of the query-key logits. To quantify this trend, we compute the Pearson correlation over $200$ images from the ImageNet-1K validation set, averaging over samples and attention heads. As expected, the correlation grows with depth, from $r \approx 0.5$ in early layers to $r \approx 0.98$ in the final layers. More details are provided in the supplementary materials. 
Notably, this correlation emerges without any pruning or ranking objective during training and is therefore learned implicitly rather than explicitly enforced.
The bias thus acts as a surrogate of column importance, providing a principled and cheap pruning signal.

\paragraph{The bias correlates with received attention.}
Finally, we examine 
\textit{property 3} from \cref{subsec:bias_definition}. Since
$g=\exp(b)$ directly reweights each column inside the softmax normalization
(\cref{eq:gate}), a positive association between the bias and the resulting attention
mass is expected by construction. However, in this analysis we investigate how strongly this correlation varies with network depth. We thus correlate $b$ with the mean post-softmax attention
received by each column across queries, using $200$ images from the ImageNet-1K
validation set and averaging over samples and attention heads. As shown in
\cref{fig:bias_attn_corr_plot}, the correlation becomes progressively stronger in deeper blocks.
Importantly, this depth-dependent behavior emerges although the model is
fine-tuned without top-$k$ selection.
Thus, while the existence of a correlation is partly induced by the formulation, its
systematic increase with depth is not imposed by the pruning mechanism and
indicates that bias implicitly becomes increasingly aligned with
column-wise attention importance.

\subsection{Bias-based attention pruning}
\label{subsec:pruning_results}

\input{tables/table_deit}

\paragraph{ImageNet Accuracy.} We now evaluate whether the learned bias can be used to reduce the computational cost of attention while preserving classification accuracy.
For each architecture, we compare three configurations. First, we consider the original model with standard self-attention. Second, we add the proposed bias term but keep all attention columns, in order to isolate the effect of the modified attention. Third, we enable bias-based pruning using the layer-wise top-\(k\) schedule $~{0.9\searrow0.1}$ described in \cref{subsec:selection_pipeline}.

As shown in \cref{tab:pruning_results}, introducing the bias alone slightly
improves accuracy for DeiT-Ti and DeiT-S, while leaving DeiT-B and
MAE-pretrained ViT-L essentially unchanged. When pruning is enabled, Top-1
accuracy decreases by approximately \(3\) points for DeiT-Ti$_b$ and \(1.3\)
points for the larger variants, relative to the corresponding original models,
while substantially reducing FLOPs and inference latency. The smallest
architecture is more sensitive to the aggressive \(0.9\searrow0.1\) schedule
because of its lower capacity and redundancy. 
Additional results in the supplementary material evaluate DeiT-Ti under multiple pruning schedules, illustrating the trade-off between accuracy, FLOPs, and latency and showing how the operating point can be adapted to different efficiency requirements.

All reported latencies are end-to-end model measurements and therefore include the overhead of bias computation, top-\(k\) selection, and index gathering.
We leverage the math attention backend to highlight the computational reduction under a controlled setting. Additional measurements in the supplementary material show consistent speedups with the efficient GPU backend and on CPU.
The efficiency gains are consistently larger for narrower models and become smaller as the embedding dimension increases. This trend aligns with the analysis in \cref{subsec:computational_gains}: for a fixed sequence length, pruning is most effective when the quadratic attention term accounts for a large fraction of the total computation, \ie when \(N\) is large relative to \(d\). Nevertheless, all four model sizes benefit from the reduced key, value, and attention dimensions, with inference speedups reaching approximately $1.47\times$.

\input{tables/table_detection}

\paragraph{Object detection and instance segmentation.}
As shown in \cref{tab:coco_transfer}, bias-based pruning transfers effectively to dense prediction. Despite the aggressive \(0.9\searrow0.1\) schedule, which retains only \(10\%\) of the keys and values in the final block, box and mask AP remain within \(0.6\) points of the baseline. This suggests that the column importance learned on ImageNet provides an effective initialization and adapts during detector fine-tuning. Although the unchanged feature pyramid and detection heads reduce the backbone-level gains, the \(1.17\times\) full-detector speedup, including post-processing, show that savings extend beyond classification.

\subsection{Combining with token merging}
\label{subsec:tome}

Bias-based pruning operates at the level of key-value columns within each attention layer, and is therefore compatible with methods that instead reduce the token sequence length. We compare our method directly with Token Merging (ToME)~\cite{bolya2023tokenmergingvitfaster}, which progressively merges similar tokens between blocks, and then combine the two approaches. We apply ToME to both DeiT-S and DeiT-S$_b$, varying the number of tokens merged per layer~$m$ to trace the accuracy-efficiency frontier. GFLOPs are measured at $1024 \times 1024$ resolution, with~$m$ rescaled proportionally to match the per-layer merging fraction used at $224 \times 224$.

\begin{figure}[t]
    \centering
    \includegraphics[width=\linewidth]{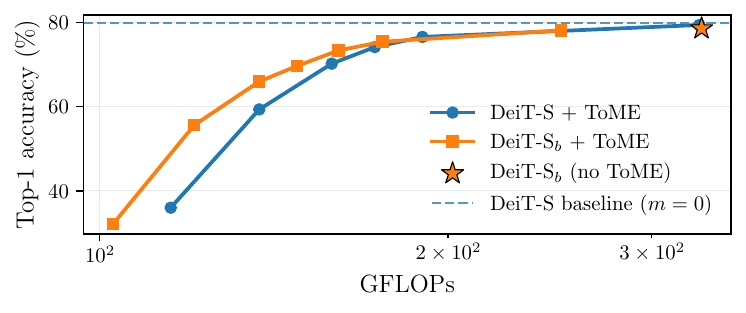}
    \caption{Accuracy-efficiency trade-off for ToME~\cite{bolya2023tokenmergingvitfaster}, our bias-based column pruning (star), and their combination on DeiT-S. The dashed line denotes the original DeiT-S accuracy.}
    \label{fig:tome}
\end{figure}

As shown in \cref{fig:tome}, ToME and our bias-based pruning (star) achieve similar accuracy at approximately the same computational budget, and both remain close to the upper bound set by the original DeiT-S. Increasing the number of tokens merged by ToME further reduces computation, but eventually causes a marked accuracy drop. Combining ToME with our column pruning provides a better trade-off in this aggressive regime: at low GFLOP budgets, DeiT-S$_b$~+~ToME retains substantially higher accuracy than applying ToME to the original DeiT-S. This confirms that the two methods exploit complementary sources of redundancy and can be effectively used together.

\input{tables/table_structured_pruning}

\subsection{Comparison with structured pruning}
\label{subsec:structured_pruning}

We additionally compare our method with structured pruning implemented using Torch-Pruning~\cite{fang2023depgraph}. Starting from the same pretrained DeiT-S, we apply global, isomorphic $\ell_2$-norm pruning with a ratio of $30\%$ and use the same fine-tuning policy as for our model. We consider two variants: standard structured pruning, which reduces channel dimensions while preserving all attention heads, and a variant that prunes $30\%$ of the heads. 

As reported in \cref{tab:structured_pruning}, standard structured pruning reaches a FLOP count comparable to our method, but yields only a marginal latency reduction.
Reducing channel dimensions does not shorten the token sequence, so the attention matrices remain quadratic in the number of tokens. Consequently, the theoretical FLOP reduction does not translate into a comparable wall-clock speedup at high resolution. Pruning attention heads reduces latency to $63.56$ ms, approximately matching our $63.08$ ms, but further lowers Top-1 accuracy. Our method therefore achieves a better accuracy-latency trade-off, reaching $78.55$\% while directly reducing the number of key-value columns during attention.

\subsection{Pruning during training and inference.}
We next study whether pruning should be applied during training, inference, or both, considering our DeiT-S$_{b}$ finetuned on ImageNet-1k. In all configurations, the model uses the proposed bias-augmented attention; we vary only whether the top-\(k\) mask is enabled in the two stages. 
The results in \cref{tab:train_test_pruning_ablation} show that inference-only pruning leads to a larger accuracy drop, since the model has not learned to compensate for the removal of low-bias columns. Applying pruning during both training and inference instead allows the network to adapt its representations to the reduced attention maps.
Disabling pruning at inference after training with pruning does not provide a clear advantage, indicating that the model adapts to operating with the selected columns rather than merely benefiting from pruning as a regularizer.

\input{tables/table_ablation_training}

%% file: tables/table_deit.tex
\begin{table}[t]
\centering
\scriptsize
\caption{Accuracy and efficiency of bias-based attention pruning on ImageNet-1k. End-to-end inference latency, including bias computation, top-\(k\) selection, index gathering, and all remaining model operations, is measured with the math attention backend on a single $1024 \times 1024$ image on one NVIDIA A40 GPU.}
\label{tab:pruning_results}
\resizebox{\linewidth}{!}{%
\begin{tabular}{lccccc}

\toprule
\multirow{2}{*}{Model}
& \multirow{2}{*}{\shortstack{Pruning\\schedule}}
& \multicolumn{2}{c}{Accuracy}
& \multirow{2}{*}{\shortstack{FLOPs\\(G)}}
& \multirow{2}{*}{\shortstack{Latency\\(ms)}} \\
\cmidrule(lr){3-4}
&
& Top-1 (\%)
& Top-5 (\%)
&
% &
&
\\
\midrule

        DeiT-Ti
        & -
        & $72.13$
        & $91.14$
        & $199.4$
        % & $0.0$
        & $47.92_{\pm 0.10}$ \\
        \midrule
        DeiT-Ti$_b$
        & \cross
        & $\mathbf{72.66}$
        & $\mathbf{91.35}$
        & $209.1$
        % & $0.0$
        & $49.04_{\pm 0.12}$ \\
        DeiT-Ti$_b$
        & \(0.9 \searrow 0.1\)
        & $69.10$
        & $89.73$
        & $\mathbf{122.1}$
        % & $0.0$
        & $\mathbf{32.52_{\pm 0.13}}$ \\

        \midrule
        \midrule
        DeiT-S
        & -
        & $79.83$
        & $94.95$
        & $485.8$
        % & $0.0$
        & $92.05_{\pm 0.14}$ \\
        \midrule
        DeiT-S$_b$
        & \cross
        & $\mathbf{80.19}$
        & $\mathbf{94.99}$
        & $505.4$
        % & $0.0$
        & $94.47_{\pm 0.16}$ \\
        DeiT-S$_b$
        & \(0.9 \searrow 0.1\)
        & $78.55$
        & $94.31$
        & $\mathbf{331.3}$
        % & $0.0$
        & $\mathbf{63.08_{\pm 0.13}}$ \\

        \midrule
        \midrule
        DeiT-B
        & -
        & $\mathbf{81.80}$
        & $\mathbf{95.59}$
        & $1319.6$
        % & $0.0$
        & $195.40_{\pm 0.33}$ \\
        \midrule
        DeiT-B$_b$
        & \cross
        & $81.79$
        & $95.47$
        & $1359.1$
        % & $0.0$
        & $200.27_{\pm 0.20}$ \\
        DeiT-B$_b$
        & \(0.9 \searrow 0.1\)
        & $80.59$
        & $94.86$
        & $\mathbf{1011.0}$
        % & $0.0$
        & $\mathbf{139.96_{\pm 0.12}}$ \\

        \midrule
        \midrule
        ViT-L (MAE)
        & -
        & $\mathbf{85.95}$
        & $\mathbf{97.57}$
        & $4131.0$
        & $556.92_{\pm 1.40}$ \\
        \midrule
        ViT-L$_b$ (MAE)
        & \cross
        & $85.85$
        & $97.55$
        & $4134.2$
        & $651.80_{\pm 0.26}$ \\
        ViT-L$_b$ (MAE)
        & \(0.9 \searrow 0.1\)
        & $84.66$
        & $97.14$
        & $\mathbf{3309.0}$
        & $\mathbf{406.98_{\pm 0.60}}$ \\
        \bottomrule
    \end{tabular}%
}

\end{table}

%% file: tables/table_detection.tex
\begin{table}[t]
    \centering
    \caption{%
        % Object detection and instance segmentation on COCO.
        Object detection and instance segmentation on COCO \texttt{val2017}. Both models use Mask R-CNN with a DeiT-S backbone and the same ViTDet-style training recipe. GFLOPs and end-to-end latency are reported for the isolated backbone and the complete detector. Latency is measured in full precision with the math attention backend on one NVIDIA A40 GPU using a single \(1024\times1024\) image.
    }
    \label{tab:coco_transfer}
    \setlength{\tabcolsep}{2.1pt}
    \resizebox{\linewidth}{!}{%
    \begin{tabular}{lccc cc cc}
        \toprule
        \multirow{2}{*}{Model}
        & \multirow{2}{*}{\shortstack{Pruning\\schedule}}
        & \multicolumn{2}{c}{COCO accuracy} & \multicolumn{2}{c}{Backbone only}
        & \multicolumn{2}{c}{Full detector} \\
        \cmidrule(lr){3-4}\cmidrule(lr){5-6}\cmidrule(lr){7-8}
        & & AP$^{\text{Box}}$ & AP$^{\text{Mask}}$ & GFLOPs & Latency (ms)
        & GFLOPs & Latency (ms) \\
        \midrule
        DeiT-S & --
            & \textbf{43.3} & \textbf{38.7}
            & 485.60 & $67.53_{\pm 0.11}$ %($1.00\times$)
            & 1171.68 & $110.37_{\pm 0.69}$ %($1.00\times$) 
            \\
        DeiT-S$_b$ & $0.9\searrow0.1$
            & 42.7 & 38.1
            & \textbf{331.17}
            & $\mathbf{51.36}_{\pm 0.09}$ % ($\mathbf{1.31\times}$)
            & \textbf{1017.25}
            & $\mathbf{94.48}_{\pm 0.52}$ %($\mathbf{1.17\times}$) 
            \\
        \bottomrule
    \end{tabular}
    }
    % \vspace{-.8em}
\end{table}

%% file: tables/table_structured_pruning.tex
\begin{table}[t]
    \centering
    \caption{Comparison with $30\%$ $\ell_2$-norm structured pruning on DeiT-S. Latency is measured with the math attention backend at $1024\times1024$ resolution on one NVIDIA A40 GPU.}
    \label{tab:structured_pruning}
    \resizebox{\linewidth}{!}{%
    \begin{tabular}{lccccc}
        \toprule
        \multirow{2}{*}{Method} & Pruning & Head & Top-1 & FLOPs & Latency  \\
        & schedule & pruning & (\%) & (G) & (ms) \\
        \midrule
        % DeiT-S & -- & $\mathbf{79.83}$ & $485.8$ & $92.05_{\pm0.14}$ \\
        DeiT-S$_{b}$ (ours) & \(0.9 \searrow 0.1\) & -- & $\mathbf{78.55}$ & $331.3$ & $\mathbf{63.08_{\pm0.13}}$ \\
        \midrule
        Structured pruning & -- & \cross & $76.80$ & $303.0$ & $88.13_{\pm0.09}$ \\
        Structured pruning & -- & \checkmark & $76.26$ & $\mathbf{301.3}$ & $63.56_{\pm0.10}$ \\
        \bottomrule
    \end{tabular}%
    }
\end{table}

%% file: tables/table_ablation_training.tex
\begin{table}[t]
\scriptsize
\centering
\caption{Effect of applying top-\(k\) pruning during training and inference on ImageNet-1k. All configurations use DeiT-S$_b$ and the \(0.9\searrow0.1\) keep-ratio schedule whenever pruning is enabled.}
\label{tab:train_test_pruning_ablation}
% \resizebox{\linewidth}{!}{%
% \begin{tabular}{cccc}
\begin{tabular*}{\columnwidth}{
    @{\extracolsep{\fill}}
    cccc
    @{}
}
\toprule
\multirow{2}{*}{\shortstack{Training\\pruning}}
& \multirow{2}{*}{\shortstack{Inference\\pruning}}
& \multicolumn{2}{c}{Accuracy} \\
\cmidrule(lr){3-4}
&
& Top-1 (\%)
& Top-5 (\%) \\
\midrule
        \cross
        & \cross
        & \(\mathbf{80.19}\)
        & \(\mathbf{94.99}\) \\

        \cross
        & \checkmark
        & \(76.90\)
        & \(93.01\) \\

        \checkmark
        & \cross
        & \(77.89\)
        & \(94.03\) \\

        \checkmark
        & \checkmark
        & \(78.55\)
        & \(94.31\) \\

        \bottomrule
    \end{tabular*}%
% }
\end{table}

%% file: sec/5_conclusion.tex
\section{Conclusion}

In this work, we introduced an input-dependent, per-head bias term to explicitly identify the query-independent components of self-attention. Our analysis shows that these components become increasingly significant as network depth increases. Our learned bias thus provides an efficient estimation of the importance of each attention column before computing the full attention matrix. 
The resulting approach is flexible by design. It adapts the selected columns to each input and attention head, and can be applied with different layer-wise pruning budgets without modifying the softmax attention operator. It is also compatible with complementary efficiency techniques, such as token merging.

Importantly, the computational savings extend beyond analytical FLOP reductions. Bias-based column pruning consistently lowers inference latency across different model sizes on real hardware, achieving speedups of up to \(1.47\times\) while retaining most of the original classification accuracy. These results indicate that exposing and exploiting query-independent token importance is a practical and broadly applicable direction for improving transformer efficiency.

%% file: sec/appendix.tex
\section*{Appendix}
\addcontentsline{toc}{section}{Appendix}
\noindent
This supplementary material is organized as follows. 
Sec.~\ref{sec_supp:details} provides training and implementation details, including a detailed pseudo-code for our method (\cref{alg:pruned_attention}). Sec.~\ref{sec_supp:ablations} reports additional pruning ablations, including the effect of different schedules on DeiT-Ti (\cref{tab:pruning_results_deit_tiny}). Sec.~\ref{sec_supp:query_free} investigates query-key-free attention, while Sec.~\ref{sec_supp:runtimes} reports latency measurements across attention backends (\cref{tab:runtime}) and hardware platforms (\cref{tab:runtime_cpu}). Sec.~\ref{sec_supp:global_vs_query} compares query-shared and query-dependent selection. Finally, Sec.~\ref{sec_supp:additional_preliminary} extends the attention and learned-bias analyses across images, architectures, and network depths.

\section{Experimental details}
\label{sec_supp:details}

\subsection{Training details}

\paragraph{ImageNet-1k (fine-tuning).}
Experiments on ImageNet ILSVRC 2012 \cite{5206848} start from the official DeiT pretrained checkpoints~\cite{deit}. We augment each self-attention module with the proposed bias term, introducing only the additional parameter matrix $W^{b}$, and, when enabled, the corresponding pruning mechanism in the forward function. All remaining model parameters are initialized from the pretrained checkpoint, after which the entire network is fine-tuned end-to-end with AdamW~\cite{adamw} for $70$ epochs, using a batch size of $1{,}024$ and an initial learning rate of $10^{-4}$, decayed with a cosine schedule without warmup with $5$ cooldown epochs. Data augmentation follows the original DeiT recipe throughout.
For ViT-L, we use the official MAE checkpoint fine-tuned on ImageNet-1K~\cite{MaskedAutoencoders2021}. After introducing the bias projection in every attention block, we fine-tune only the attention modules for $30$ epochs on four NVIDIA A40 GPUs. We use AdamW, an effective batch size of $1{,}024$, layer-wise learning-rate decay of $0.75$, weight decay of $0.05$, and the MAE fine-tuning augmentations. We train both the bias-without-pruning and $0.9\searrow0.1$ pruning configurations with the same recipe.

\paragraph{COCO object detection and instance segmentation.}
We evaluate Mask R-CNN~\cite{he2017maskrcnn} with a DeiT-S backbone on COCO~\cite{lin2014coco} \texttt{train2017} and report box and mask AP on \texttt{val2017}. Following the ViTDet~\cite{li2022vitdet} protocol, a Simple Feature Pyramid converts the single-scale DeiT representation into the hierarchy used by the detection and segmentation heads. Training uses large-scale jittering with a fixed \(1024\times1024\) image size, AdamW, mixed precision, and layer-wise learning-rate decay. Both models are fine-tuned end-to-end for 100 epochs on 4 NVIDIA A40 GPUs with a global batch size of 16. The baseline is initialized from the official supervised DeiT-S ImageNet-1K checkpoint, whereas the pruned model uses the corresponding bias-augmented checkpoint and the \(0.9\searrow0.1\) schedule. During training, top-\(k\) masking is applied to the full attention logits; at validation time, discarded keys and values are removed before attention. All remaining architectural and optimization choices are identical.

Efficiency is measured in full precision on one NVIDIA A40 GPU using the \texttt{math} attention backend and a single \(1024\times1024\) validation image. Full-detector measurements include the Simple Feature Pyramid, region proposal network, RoI heads, and prediction post-processing. The backbone FLOPs differ slightly from the classification results because the dense-prediction backbone omits the classification token and head.

\paragraph{ImageNet100 (training from scratch).}
Models evaluated on ImageNet100 (a $100$-class subset of ImageNet) are trained
from scratch for $300$ epochs using a recipe inspired by DeiT~\cite{deit}, with
some simplifications.
We use AdamW with an initial learning rate of $10^{-3}$ decayed with a cosine
schedule and a linear warmup of $5$ epochs (warmup decay factor $0.033$).
The batch size is $1\,024$ and weight decay is set to $0.1$.
Gradient norms are clipped to $1.0$.
For regularization and data augmentation, we use label smoothing ($\varepsilon =
0.1$), RandAugment with $2$ operations at magnitude $9$, Mixup with intensity
parameter $\alpha = 0.2$ (i.e.\ sampled from $\mathrm{Beta}(0.2, 0.2)$), and
Cutmix with $\alpha = 1.0$.
The complete set of hyper-parameters is reported in
\cref{tab:training_recipe}.

\begin{table}[t]
    \centering
    \footnotesize
    \caption{%
        Training recipe used for the ImageNet100 experiments.
        RandAugment ``2/9'' denotes 2 operations at magnitude 9, following the
        \texttt{torchvision} implementation.
        Mixup ($\alpha=0.2$) and Cutmix ($\alpha=1.0$) intensities follow the
        PyTorch convention: $\alpha$ is the parameter of the symmetric Beta
        distribution from which the mixing coefficient is sampled.
    }
    \label{tab:training_recipe}
    \begin{tabular*}{.7\columnwidth}{
            @{\extracolsep{\fill}}
            lr
            @{}
        }
        \toprule
        Hyperparameter &  \\
        \midrule
        Epochs              & 300   \\
        Batch size          & 1\,024 \\
        Optimizer           & AdamW \\
        Learning rate       & $10^{-3}$ \\
        LR decay            & cosine \\
        Warmup epochs       & 5     \\
        Warmup decay        & 0.033 \\
        Weight decay        & 0.1   \\
        Gradient clipping   & 1.0   \\
        \midrule
        Label smoothing     & 0.1   \\
        \midrule
        RandAugment         & 2/9   \\
        Mixup               & 0.2   \\
        Cutmix              & 1.0   \\
        \bottomrule
    \end{tabular*}
\end{table}

\input{sec/pseudo_code}

\subsection{Implementation details}

\paragraph{Codebase.}
Our implementation builds on the official DeiT~\cite{deit} and MAE~\cite{MaskedAutoencoders2021} codebases and the
\texttt{timm} library~\cite{timm}. We modify only the self-attention module to
introduce the input-dependent bias and the bias-based column pruning, while the
data pipeline, optimizer, and training loop follow DeiT unchanged. The released
code preserves the original Apache~2.0 license and attribution. Bias computation
and top-$k$ selection are implemented with standard PyTorch tensor operations and
require no custom CUDA kernels or specialized sparse-attention libraries.

\paragraph{Bias projection.}
For each attention head $i$, the input-dependent bias $b_i = X W^b_i$ is
computed by a single linear projection $W^b_i \in \mathbb{R}^{d \times 1}$
applied to the token sequence $X \in \mathbb{R}^{N \times d}$. The complete pseudo-code for the updated forward pass is reported in \cref{alg:pruned_attention}.

\paragraph{Top-$k$ selection and actual speedups.}
Bias values are computed before the attention scores, so the top-$k$ indices
are available before forming $QK^\top$. Each attention head selects the same
number $k$ of tokens, although the selected indices may differ across heads.
In our implementation, the key and value projections are first computed for all
$N$ tokens, after which the $k$ selected keys and values are gathered separately
for each head according to the retaining set $\mathcal{K}$. Because the retained cardinality is identical across heads, the
resulting tensors have uniform dense shapes and can be processed efficiently
using standard batched matrix multiplications.

Consequently, the full $N\times N$ attention matrix is never materialized.
Instead, each head operates on a score matrix of size $N\times \vert \mathcal{K} \vert$, and both
scaled dot-product attention and value aggregation are performed only over the
selected columns. 

These operations are
implemented using standard dense batched matrix multiplications and indexing
operations, requiring neither custom CUDA kernels nor sparse-attention
libraries. The resulting speedup therefore comes from reducing the quadratic
attention-score.

\paragraph{Layer-wise keep-ratio schedule.}
For a network with $L$ layers, we define a linearly decreasing keep ratio
\[
  \rho_\ell = \rho_{\mathrm{high}}
  - (\rho_{\mathrm{high}} - \rho_{\mathrm{low}})\,\frac{\ell-1}{L-1},
  \qquad k_\ell = \lfloor \rho_\ell N \rceil,
\]
where $\ell = 1,\ldots,L$.
For the schedule $0.9 \searrow 0.1$ on the 12-layer DeiT architectures, this
yields the per-layer keep ratios
0.900, 0.827, 0.755, 0.682, 0.609, 0.536, 0.464, 0.391, 0.318, 0.245, 0.173,
0.100.

\paragraph{Pruning during training.}
When pruning is enabled at training time, the top-$k$ mask is applied in the
forward pass as a hard selection: discarded columns receive no gradient, but the
bias weights $W^b_i$ are still updated through the retained columns.
This allows the model to adapt its internal representations to the reduced
attention maps and yields better accuracy than applying pruning only at
inference (see \cref{tab:train_test_pruning_ablation} in the main paper).

\input{tables/table_deit-tiny}

\section{Additional pruning ablations}
\label{sec_supp:ablations}

\subsection{Effect of the pruning schedule on DeiT-Ti}
\label{sec:supp_tiny_schedule}

\Cref{tab:pruning_results} of the main paper compares all architectures using the same aggressive
\(0.9\searrow0.1\) pruning schedule. This operating point can be particularly
challenging for small models such as DeiT-Ti.

Our method, however, allows the layer-wise pruning schedule to be adapted to
the desired accuracy-efficiency trade-off. In
\cref{tab:pruning_results_deit_tiny}, we retain \(90\%\) of the columns in the
first layer and vary the final-layer keep ratio from \(0.5\) to \(0.1\). A more
conservative \(0.9\searrow0.5\) schedule preserves most of the original
accuracy while still reducing FLOPs and latency. As expected, lower final keep ratios
progressively increase the computational savings at the cost of larger
accuracy degradation. 

\input{tables/table_ablation_column}

\input{tables/table_ablation_prob}

\input{tables/table_ablation_topk}

\subsection{Column-selection strategy}
We compare the layer-wise top-\(k\) schedule used in our main experiments with a fixed keep ratio, and a probability-threshold rule. Given the normalized scores \(p_k = \frac{\exp(b_k)}{\sum_j \exp(b_j)}\), probability thresholding sorts columns by decreasing \(p_k\) and retains the smallest set whose cumulative probability mass reaches a threshold \(\tau\). Unlike top-\(k\), this criterion allows the number of retained columns to vary across inputs and heads. However, the resulting variable tensor shapes do not directly support efficient dense batched computation. Experiments are conducted on ViT-S$_b$ trained on ImageNet100.

As shown in \cref{tab:selection_ablation}, at an average pruning ratio of approximately \(50\%\), the layer-wise \(0.9\searrow0.1\) schedule outperforms both a fixed \(0.5\) keep ratio and probability thresholding. This confirms that applying the same pruning rate at every depth is sub-optimal. Probability thresholding offers greater flexibility across samples and heads, but top-\(k\) produces fixed tensor shapes and therefore enables the actual reduction of the key, value, and attention dimensions.

\subsection{Probability-threshold pruning}
\label{subsec:supp_prob_threshold}

The preceding analysis reports probability-threshold results only at a single operating
point ($\tau = 0.8$, corresponding to roughly 50\% pruning ratio) for a direct
comparison with the fixed and layer-wise top-$k$ schedules.
\cref{tab:score_threshold_imagenet100} extends this analysis by sweeping $\tau$ across a
wider range.

Given the normalized bias scores
$p_k = \exp(b_k)/\sum_j \exp(b_j)$,
probability thresholding retains the smallest set of tokens whose cumulative
probability mass reaches $\tau$.
This makes the number of retained columns input- and head-dependent, which
in principle allows a finer-grained adaptation to each sample.
However, the resulting variable tensor shapes prevent direct use of dense
batched matrix multiplications and therefore limit practical wall-clock
speedups to those achievable through masked (padded) implementations.

At comparable pruning ratios, the layer-wise top-$k$ schedule consistently
outperforms probability thresholding (\cref{tab:selection_ablation}), confirming that
the fixed-shape constraint of top-$k$ comes at no accuracy cost in practice.

\subsection{Fixed top-$k$ schedules}
\label{subsec:supp_topk}

We additionally evaluate uniform keep ratios, where the same fraction of
columns is retained in every layer, at several pruning levels, and compare
them with the layer-wise schedule adopted in the main paper.
Results are reported in \cref{tab:topk_imagenet100}.

Across all tested keep ratios, the layer-wise $0.9 \searrow 0.1$ schedule
outperforms the corresponding uniform baseline at a comparable average pruning
ratio.
This confirms that allocating a larger
attention budget to early layers, where query-dependent patterns are most
prominent, and progressively increasing the pruning ratio with depth is
consistently more effective than applying a uniform reduction throughout the
network.

\input{tables/table_bias_only}

\section{Query-key-free attention in the last layers}
\label{sec_supp:query_free}

We evaluate the architectural variant discussed in \cref{subsec:prelim_experiments} of the main
paper, in which the query-key term
$Q_iK_i^{\top}/\sqrt{d_k}$ is removed from the last transformer blocks and
attention is determined solely by the broadcast bias $\mathbf{1}b_i^\top$.
We conduct this analysis on ViT-S$_b$ trained from scratch on ImageNet100. The
network has $12$ transformer blocks, and we progressively replace full-attention
logits with the query-key-free formulation in the final $4$, $6$, or $8$ blocks.

As reported in \cref{tab:bias_only_imagenet100}, using query-key-free attention in
the last four blocks preserves and slightly improves Top-1 accuracy over the
original ViT-S, indicating that query-key interactions become largely
redundant near the end of the network. Extending the replacement to the last
six or eight blocks produces a moderate degradation, showing that
query-dependent interactions remain useful at intermediate depths. Overall,
the result further supports our interpretation that the query-independent
component becomes increasingly dominant in deeper layers.

\begin{table*}[t]
    \centering
    \caption{%
        Inference latency (ms) of the original and bias-pruned models using the
        \texttt{math} and \texttt{efficient} attention backends. Measurements
        use a single $1024\!\times\!1024$ image on one NVIDIA A40 GPU. Values in
        parentheses report the speedup of the pruned model over its corresponding
        original model.
    }
    \label{tab:runtime}
    \setlength{\tabcolsep}{7pt}
    \begin{tabular*}{\textwidth}{
            @{\extracolsep{\fill}}
            l cc  cc
            @{}
        }
        \toprule
        & \multicolumn{2}{c}{\texttt{math} backend} &
          \multicolumn{2}{c}{\texttt{efficient} backend} \\
        \cmidrule(lr){2-3}
        \cmidrule(lr){4-5}
        Model & Original & Bias-pruned & Original & Bias-pruned \\
        \midrule
        DeiT-Ti & $47.92_{\pm0.10}$ & $\mathbf{32.52_{\pm0.13}}\;(1.47\times)$
                & $33.60_{\pm0.30}$ & $\mathbf{20.83_{\pm0.20}}\;(1.61\times)$ \\
        DeiT-S  & $92.05_{\pm0.14}$ & $\mathbf{63.08_{\pm0.13}}\;(1.46\times)$
                & $63.22_{\pm0.09}$ & $\mathbf{39.88_{\pm0.09}}\;(1.59\times)$ \\
        DeiT-B  & $195.40_{\pm0.33}$ & $\mathbf{139.96_{\pm0.12}}\;(1.40\times)$
                & $140.59_{\pm0.16}$ & $\mathbf{93.61_{\pm0.16}}\;(1.50\times)$ \\
        ViT-L (MAE) & $556.92_{\pm1.40}$ & $\mathbf{406.98_{\pm0.60}}\;(1.37\times)$
                & $417.46_{\pm0.79}$ & $\mathbf{288.55_{\pm0.37}}\;(1.45\times)$ \\
        \bottomrule
    \end{tabular*}
\end{table*}

\begin{table}[t]
\footnotesize
    \centering
    \caption{%
        CPU inference latency (ms) of the original and bias-pruned models.
        Measurements use one $1024\!\times\!1024$ image on an AMD EPYC 7402
        processor with 12 CPU threads and the \texttt{math} attention backend.
        We report the mean and standard deviation over $20$ runs after $5$ warm-up
        iterations. Parentheses indicate the speedup over the corresponding
        original model.
    }
    \label{tab:runtime_cpu}
    \setlength{\tabcolsep}{5pt}
    \begin{tabular*}{\columnwidth}{
            @{\extracolsep{\fill}}
            lrr
            @{}
        }
        \toprule
        Model & Original & Bias-pruned \\
        \midrule
        DeiT-Ti
            & $1,886.12_{\pm53.98}$
            & $\mathbf{966.00_{\pm86.12}}\;(1.95\times)$ \\
        DeiT-S
            & $2,325.13_{\pm49.94}$
            & $\mathbf{1,369.32_{\pm75.35}}\;(1.70\times)$ \\
        DeiT-B
            & $ 6,770.29_{\pm70.85}$
            & $\mathbf{3,891.03_{\pm252.29}}\;(1.74\times)$ \\
        ViT-L (MAE)
            & $21,187.74_{\pm440.54}$
            & $\mathbf{11,807.39{\pm166.21}}\;(1.79\times)$ \\
        \bottomrule
    \end{tabular*}
\end{table}

\section{Runtime measurements and backend considerations}
\label{sec_supp:runtimes}
\label{sec:supp_runtime}

\paragraph{Backend dependence.}
Modern PyTorch scaled dot-product attention is dispatched to one of three
backends: \texttt{flash}, \texttt{efficient}, or \texttt{math}.
The main paper reports measurements with the \texttt{math} backend, whose
straightforward matrix multiplications make the relationship between FLOPs and
latency easier to interpret. Here we additionally evaluate the
\texttt{efficient} backend, which uses optimized kernels and represents a more practical inference setting. Our implementation relies on dense
fixed-shape tensors and standard PyTorch operations, and therefore supports
both backends without custom kernels.

\paragraph{Comparison across backends.}
\cref{tab:runtime} compares the original and bias-pruned configurations for
DeiT-Ti/S/B and MAE-pretrained ViT-L under both backends.
Measurements are taken on a single NVIDIA A40 GPU at $1024 \times 1024$
resolution, averaged over 200 forward passes after a warm-up of 50 passes.
Column pruning consistently reduces latency regardless of the selected backend.
With \texttt{math}, speedups range from $1.37\times$ to $1.47\times$; with
\texttt{efficient}, they range from $1.45\times$ to $1.61\times$. The gains
under the optimized backend confirm that our dense implementation is compatible
with efficient attention kernels and that its acceleration is not restricted
to the controlled \texttt{math} setting used in the main paper.

\paragraph{CPU measurements.}
We additionally evaluate CPU inference latency on an AMD EPYC 7402 processor.
Measurements use a batch size of one and a single
$1024\times1024$ image with the \texttt{math} attention backend. We report the
mean and standard deviation over $20$ forward passes after $5$ warm-up iterations.

As shown in \cref{tab:runtime_cpu}, bias-based column pruning provides a
substantial CPU acceleration, reaching speedups up to $1.95\times$ for DeiT-Ti. These gains show that the reduced key, value, and
attention dimensions also translate into lower latency outside GPU-specific
attention kernels.

\section{Global vs. query-dependent selection}
\label{sec_supp:global_vs_query}

\Cref{fig:attention_sparsity} of the main paper qualitatively shows that the attention maps of a
pretrained DeiT-S progressively exhibit vertical structures as depth increases (\ie
small number of key columns receive attention from many different queries).
These patterns suggest that, especially in deeper layers, it may be possible
to identify a common subset of key tokens that is important to most queries
and remove the remaining columns altogether. However, \cref{fig:attention_sparsity} alone does not
quantify how much attention mass such a shared selection would preserve, nor
how much would be lost compared with a finer-grained selection tailored to
each query.

To measure this trade-off, we compare two sparsification strategies under the
same pruning-column budget. The first assigns each key token a single importance score. This \textit{query-shared} selection allows
entire key-value columns to be removed and is therefore the form of sparsity
required by our method to reduce the attention matrix from $N\times N$ to
$N\times k$.
The second is a \textit{query-dependent} strategy, which retains the largest attention
entries in each query row independently. This method can better preserve query-specific interactions, but generally selects different key tokens for different queries and therefore
does not correspond to removing complete columns. We treat this finer-grained
strategy as an oracle at the same sparsity budget and measure how closely the
query-shared selection approaches it as network depth increases. For this purpose, we next define the \emph{global top-$k$} set shared across
queries and the corresponding \emph{query-dependent top-$k$} sets.

Notably, this analysis is conducted on the original pretrained DeiT-S,
without introducing our bias term or fine-tuning the model. It therefore
quantifies a property that emerges spontaneously in standard self-attention.

\begin{figure*}[t]
    \centering
    \begin{subfigure}[t]{.49\textwidth}
        \centering
        \includegraphics[width=\linewidth]{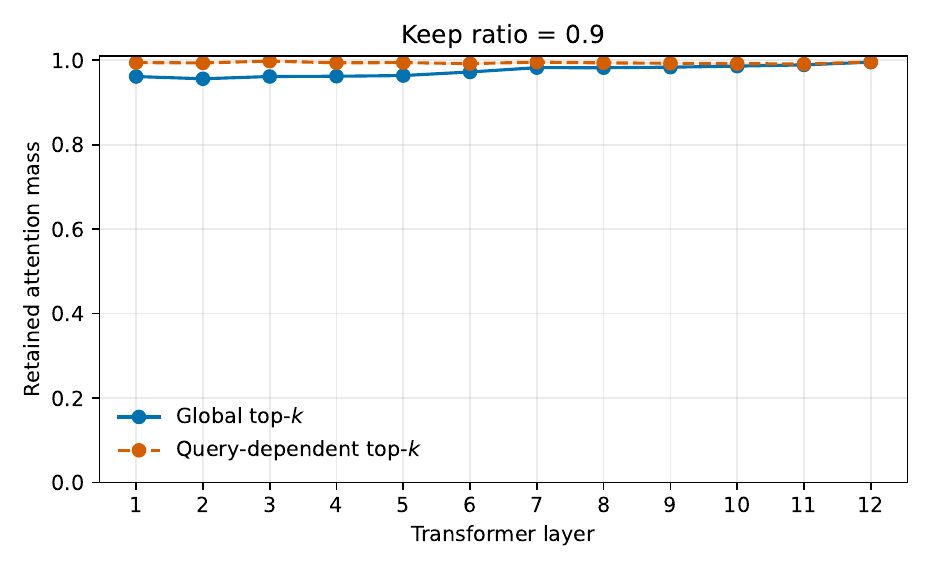}
        \caption{Keep ratio $\rho=0.9$.}
        \label{fig:supp_global_query_09}
    \end{subfigure}
    \hfill
    \begin{subfigure}[t]{.49\textwidth}
        \centering
        \includegraphics[width=\linewidth]{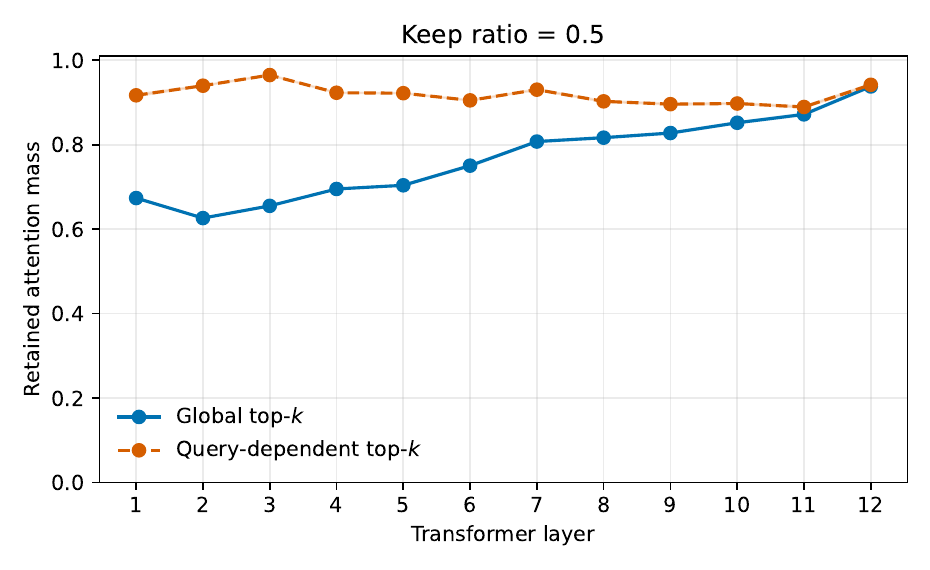}
        \caption{Keep ratio $\rho=0.5$.}
        \label{fig:supp_global_query_05}
    \end{subfigure}

    \medskip
    \begin{subfigure}[t]{.49\textwidth}
        \centering
        \includegraphics[width=\linewidth]{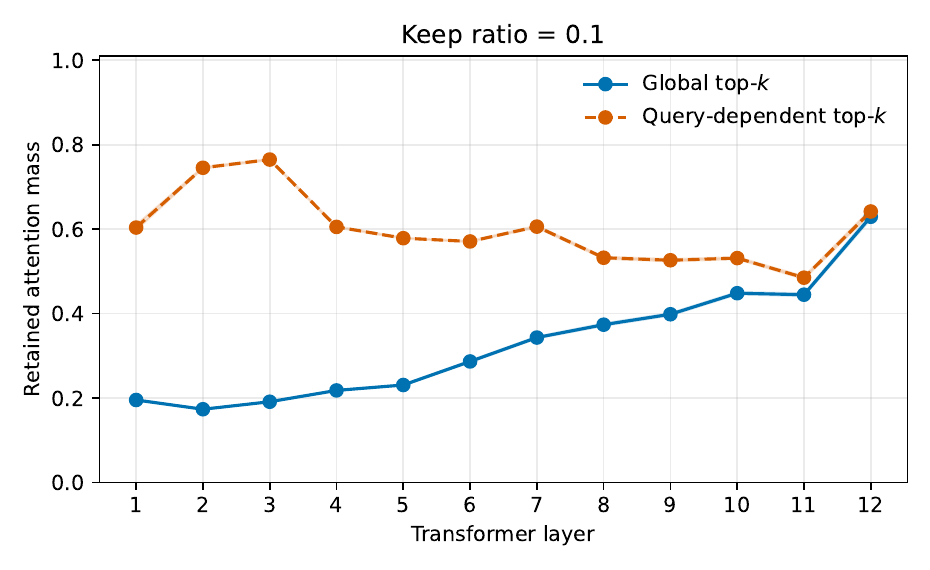}
        \caption{Keep ratio $\rho=0.1$.}
        \label{fig:supp_global_query_01}
    \end{subfigure}
    \hfill
    \begin{subfigure}[t]{.49\textwidth}
        \centering
        \includegraphics[width=\linewidth]{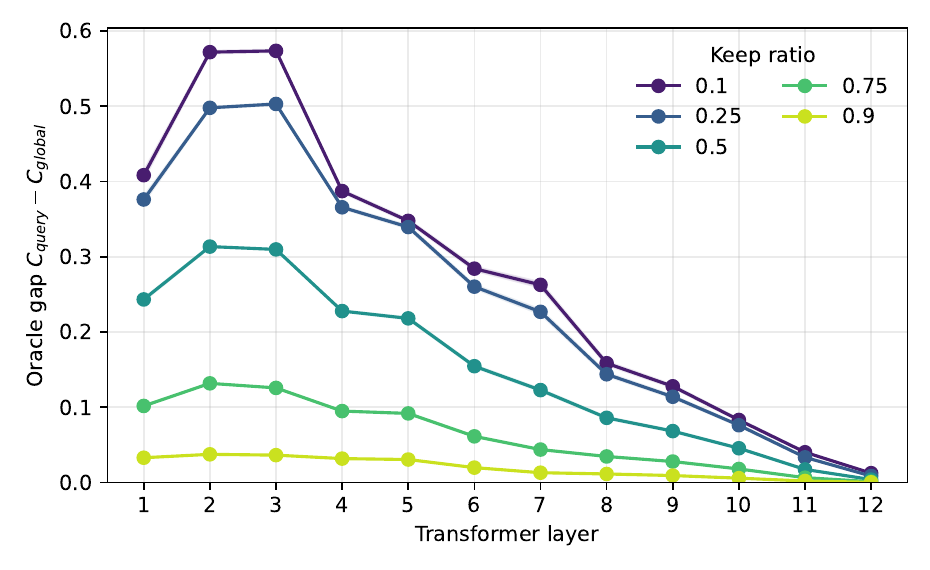}
        \caption{Oracle gap across keep ratios.}
        \label{fig:supp_global_query_gap}
    \end{subfigure}
    \caption{%
        Global versus query-dependent column selection in the original
        pretrained DeiT-S, averaged over all attention heads and $200$
        ImageNet-1K validation images. \textbf{(a--c)} Retained attention mass
        at different keep ratios. \textbf{(d)} Oracle gap
        $\Delta=C_{\mathrm{query}}-C_{\mathrm{global}}$ across all evaluated
        keep ratios.
    }
    \label{fig:supp_global_query_topk}
\end{figure*}
\paragraph{Global and query-dependent top-$k$.}
Let
\begin{equation}
    A^{(s,\ell,h)}
    = \operatorname{softmax}\!\left(
        Q^{(s,\ell,h)}K^{(s,\ell,h)\top}/\sqrt{d_k}
      \right)
    \in \mathbb{R}^{N\times N}
\end{equation}
denote the post-softmax attention matrix for image $s$, layer $\ell$, and
head $h$. As in \cref{fig:attention_sparsity} of the main paper, we remove the class-token row and
column and renormalize every remaining row, so that $N=196$ denotes the number
of patch tokens and $\sum_{j=1}^{N} A_{qj}^{(s,\ell,h)}=1$ for every patch
query $q$. For readability, we omit the superscripts $(s,\ell,h)$ below.

For a keep ratio $\rho$, we retain
$m_\rho=\max(1,\lfloor \rho N\rceil)$ columns. We first define the mean
attention received by column $j$ across all queries as
\begin{equation}
    \bar{a}_j = \frac{1}{N}\sum_{q=1}^{N} A_{qj}.
\end{equation}
The \emph{global top-$k$} set is then
\begin{equation}
    \mathcal{G}_\rho = \operatorname{TopK}(\bar{a},m_\rho),
\end{equation}
and is shared by every query within the image and attention head. Its retained
attention mass is
\begin{equation}
    C_{\mathrm{global}}(\rho)
    = \frac{1}{N}\sum_{q=1}^{N}
      \sum_{j\in\mathcal{G}_\rho} A_{qj}.
    \label{eq:supp_c_global}
\end{equation}
Here, ``global'' means global across queries: $\mathcal{G}_\rho$ is still
input-, layer-, and head-dependent.

For comparison, the \emph{query-dependent top-$k$} independently selects the
best columns for each query,
\begin{equation}
    \mathcal{Q}_{q,\rho}
    = \operatorname{TopK}(A_{q,:},m_\rho),
\end{equation}
with retained mass
\begin{equation}
    C_{\mathrm{query}}(\rho)
    = \frac{1}{N}\sum_{q=1}^{N}
      \sum_{j\in\mathcal{Q}_{q,\rho}} A_{qj}.
    \label{eq:supp_c_query}
\end{equation}
Because it may choose a different set for every row,
$C_{\mathrm{query}}$ is an oracle upper bound at the same column budget, and
$C_{\mathrm{query}}\geq C_{\mathrm{global}}$ by construction. We therefore
measure the cost of enforcing a query-shared selection through the oracle gap
\begin{equation}
    \Delta(\rho)
    = C_{\mathrm{query}}(\rho)-C_{\mathrm{global}}(\rho).
    \label{eq:supp_oracle_gap}
\end{equation}
We compute these quantities for every layer and attention head on a random
subset of $200$ ImageNet-1K validation images. Heads are first averaged
within each image, after which we report the mean across images.

\paragraph{Results and relation to depth.}
With mild pruning ($\rho=0.9$), both strategies preserve almost all attention
mass throughout the network and their gap is small
(\cref{fig:supp_global_query_09}). As pruning becomes more aggressive, the gap
in the early layers increases substantially
(\cref{fig:supp_global_query_05,fig:supp_global_query_01}), revealing the
importance of query-specific interactions at shallow depth. Nevertheless (after peaking in the early blocks), the oracle gap consistently decreases with
depth for every keep ratio and approaches zero in the final layers
(\cref{fig:supp_global_query_gap}). 

Thus, the important-column ranking becomes
increasingly shared across queries, quantitatively supporting the vertical
structures in \cref{fig:attention_sparsity} of the main paper. This result further supports our
$0.9\searrow0.1$ schedule, which preserves more columns where query-specific
structure is strongest and prunes progressively with depth. Moreover, it identifies
a failure regime of our strategy, when aggressive column pruning in the early layers is adopted.

\section{Additional analysis of the learned bias}
\label{sec_supp:additional_preliminary}

\subsection{Additional attention visualizations}

\paragraph{Setup.}
In the main paper, attention-map visualizations (\cref{fig:attention_sparsity}) were restricted to a
single validation image and a single attention head of DeiT-S for readability.
Here we extend the analysis along four additional axes: (i) \emph{multiple
images}, to assess whether the qualitative emergence of column sparsity generalizes across images; (ii) \emph{multiple layers} spanning the network depth, rather than only the three representative blocks shown in the main paper; (iii) multiple heads and (iv) \emph{multiple models}, namely
DeiT-Ti, DeiT-S, DeiT-B, and MAE-pretrained ViT-L, to verify that the effect is not
specific to a particular model size.

\begin{figure*}[p]
    \centering
    \includegraphics[width=\linewidth]%
        {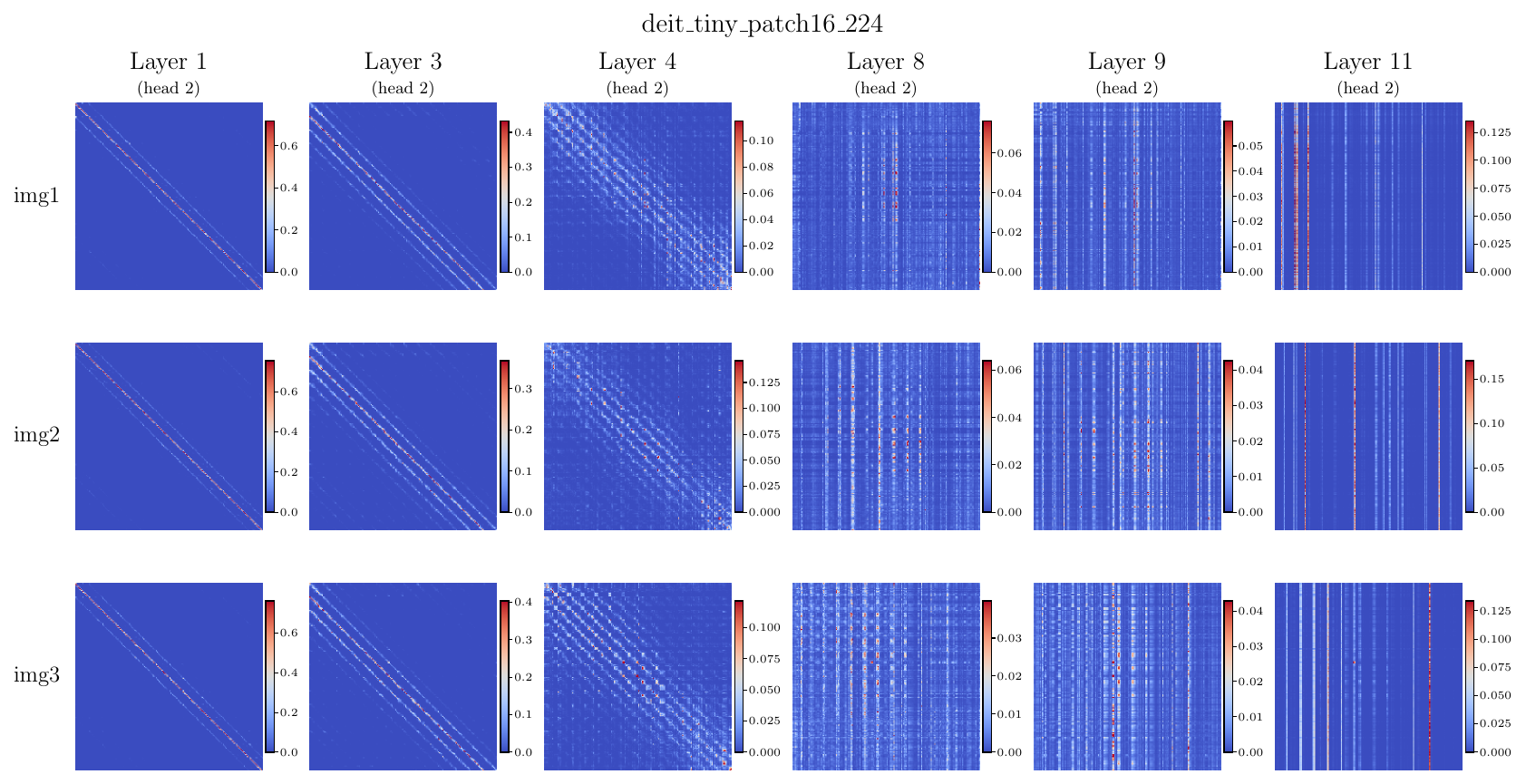}
    \caption{%
        Attention maps for DeiT-Ti at layers $\{1,3,4,8,9,11\}$ (columns),
        head~2, across several ImageNet validation images (rows).
        In all cases the attention evolves from a diagonal, locally-concentrated
        pattern in early layers to strongly column-sparse maps in the final
        blocks, where a small number of tokens attract the attention of virtually
        all queries.
    }
    \label{fig:attn_tiny}
\end{figure*}

\begin{figure*}[p]
    \centering
    \includegraphics[width=\linewidth]%
        {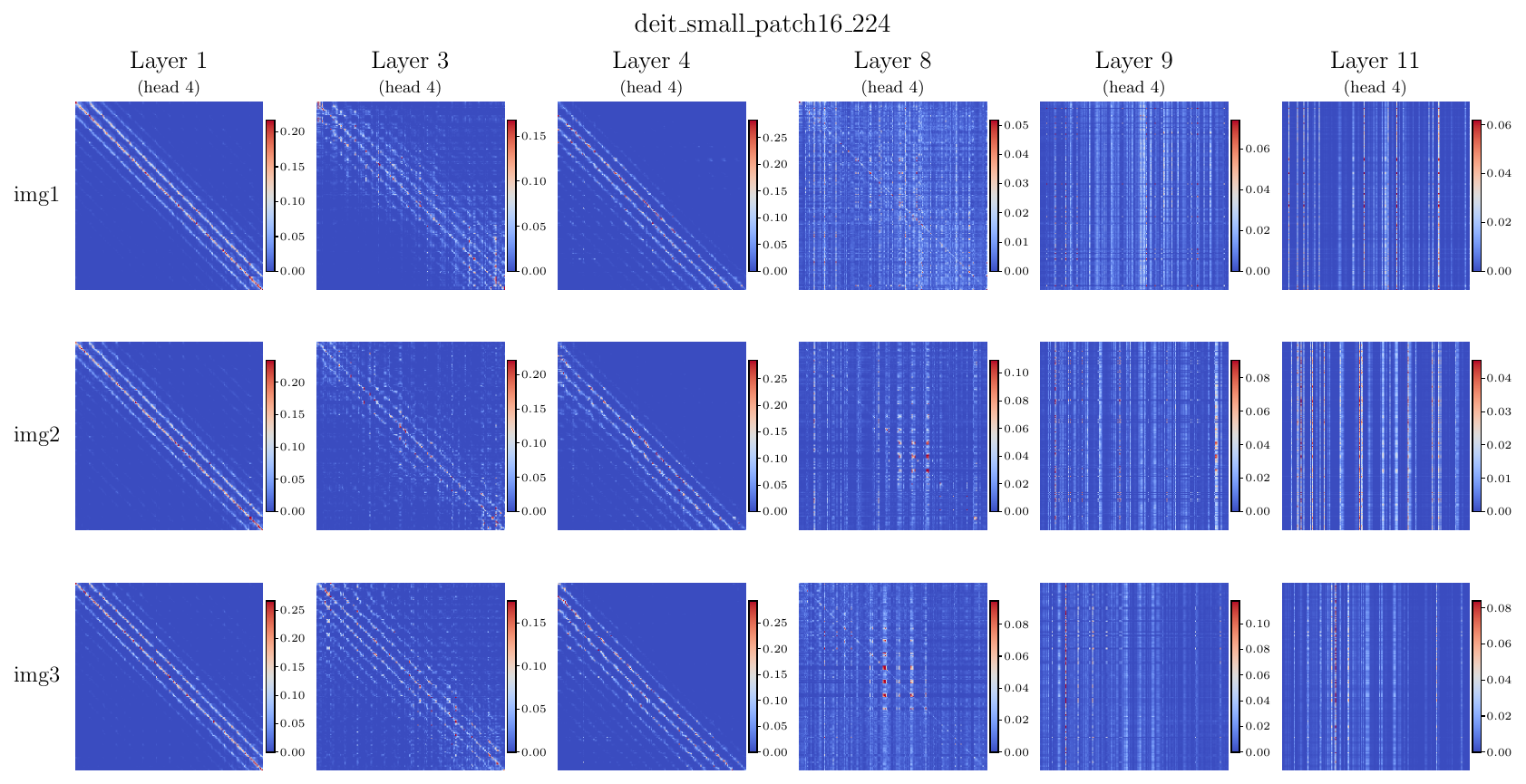}
    \caption{%
        Attention maps for DeiT-S at layers $\{1,3,4,8,9,11\}$ (columns),
        head~4, across several ImageNet validation images (rows).
        The transition from diagonal to vertical attention structure is visible
        and consistent across all images, confirming that column sparsity is
        a systematic property of the attention mechanism rather than an
        image-specific artifact.
    }
    \label{fig:attn_small}
\end{figure*}

\begin{figure*}[p]
    \centering
    \includegraphics[width=\linewidth]%
        {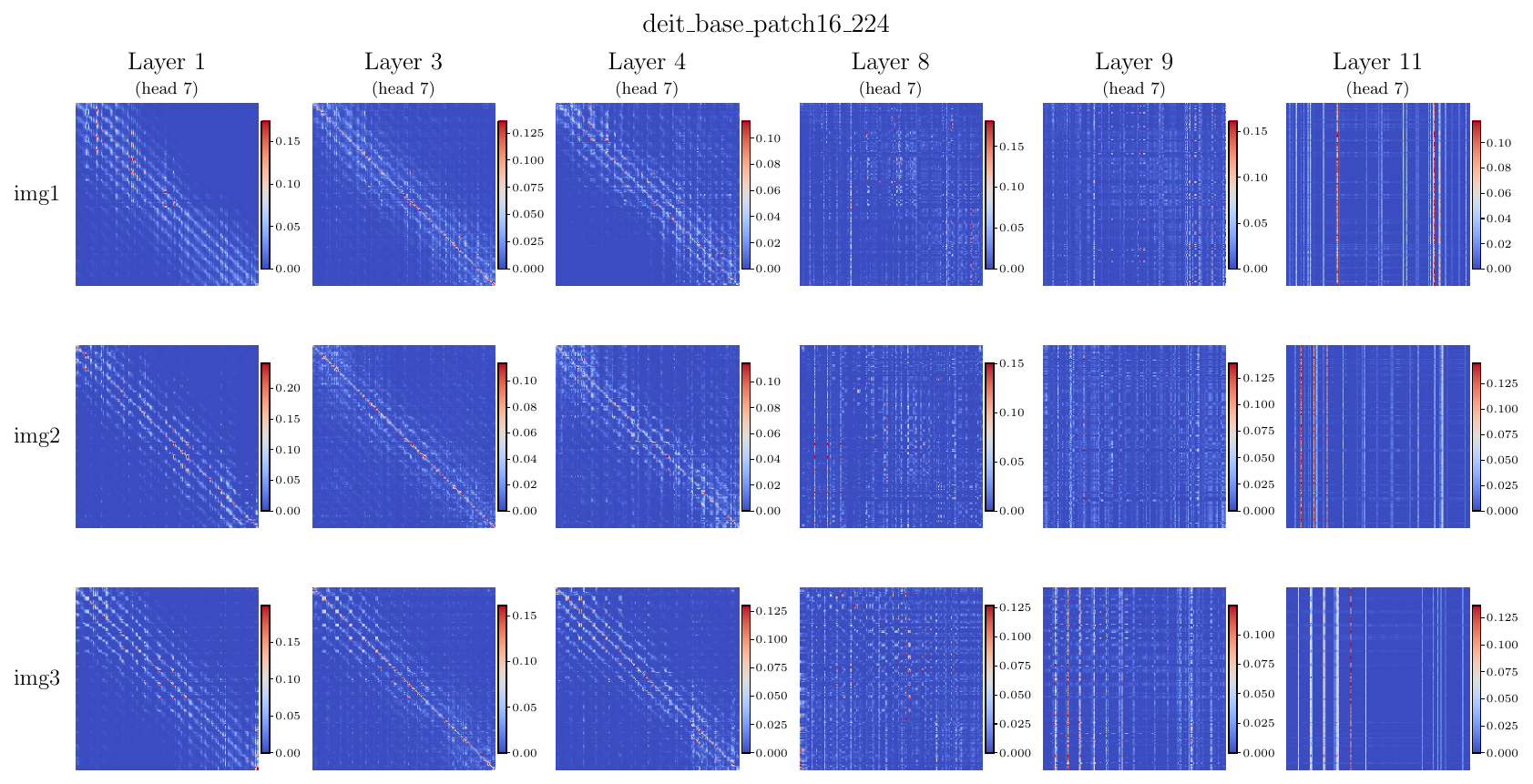}
    \caption{%
        Attention maps for DeiT-B at layers $\{1,3,4,8,9,11\}$ (columns),
        head~7, across several ImageNet validation images (rows).
        The same depth-wise evolution observed in DeiT-Ti and DeiT-S is
        reproduced here, showing that column sparsity is not specific to a
        particular model width.
    }
    \label{fig:attn_base}
\end{figure*}

\paragraph{Results across images and layers.}
\cref{fig:attn_tiny,fig:attn_small,fig:attn_base} report attention maps for
DeiT-Ti, DeiT-S, and DeiT-B, respectively, at layers
$\{1, 3, 4, 8, 9, 11\}$ across several validation images.
In all three models and across all tested images, the qualitative evolution with
depth is consistent: early layers display a pronounced diagonal structure,
indicating that each token primarily attends to itself and its spatial
neighbours; as depth increases, vertical structures progressively emerge and
eventually dominate the attention maps in the final layers.
This pattern holds across different images belonging to the ImageNet validation set,
confirming that column sparsity is a systematic property of the attention
mechanism rather than an image-specific artifact.
Importantly, the same phenomenon is visible in all three architectures, with
the transition from diagonal to vertical structure occurring at comparable
relative depths regardless of model width.

\begin{figure*}[p]
    \centering
    \includegraphics[width=\linewidth]%
        {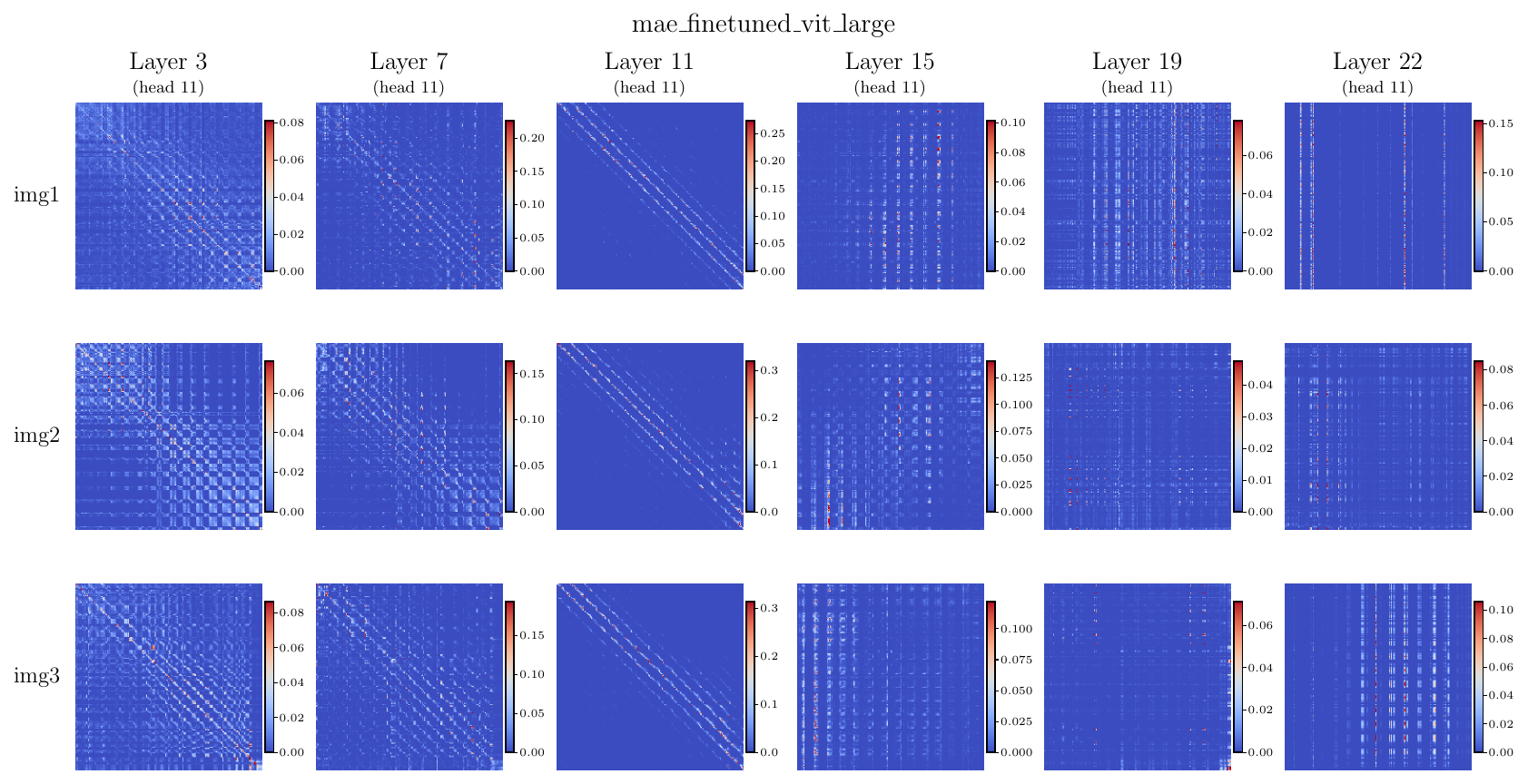}
    \caption{%
        Attention maps for the MAE-pretrained ViT-L at layers
        $\{3,7,11,15,19,22\}$ (columns), head~11, across several ImageNet
        validation images (rows). Despite its greater depth and different
        pretraining strategy, ViT-L exhibits the same transition from
        predominantly diagonal attention in early blocks to increasingly
        column-sparse patterns in deeper blocks.
    }
    \label{fig:attn_vit_large}
\end{figure*}

As shown in \cref{fig:attn_vit_large}, we additionally inspect the 24-block MAE-pretrained ViT-L at layers
$\{3,7,11,15,19,22\}$ and across its attention heads. The same transition is
preserved at this larger depth and scale: early attention is predominantly
diagonal, whereas a small set of columns becomes dominant in the final blocks.

\begin{figure}[t]
    \centering
    \includegraphics[width=.8\linewidth]%
{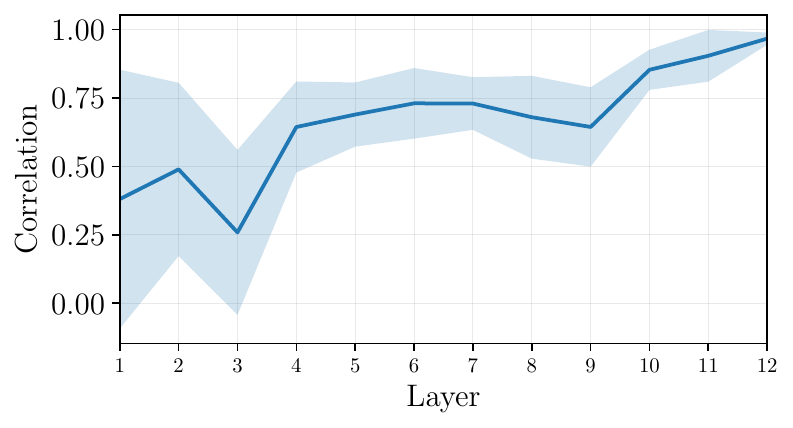}
    \caption{%
        Layer-wise Pearson correlation between the learned bias and the
        pre-softmax per-column mean $\operatorname{mean}_q(QK^\top)$.
        The correlation grows steadily with depth, reaching $r \approx 0.98$
        in the final layer, confirming that the bias serves as an effective
        and cheap surrogate of column importance.
    }
    \label{fig:pre-soft}
\end{figure}

\paragraph{Head selection.} Each attention-map in \cref{fig:attn_tiny,fig:attn_small,fig:attn_base,fig:attn_vit_large} reports one representative head for each architecture: head 2 for DeiT-Ti, head 4 for DeiT-S, head 7 for DeiT-B, and head 11 for ViT-L. Although the selected heads differ across architectures, all exhibit the same qualitative depth-wise transition from predominantly diagonal attention to increasingly column-sparse patterns. Individual heads may differ in the location and sharpness of the dominant columns but present the same patterns.

\begin{figure*}[p]
    \centering
    \includegraphics[width=\linewidth]%
        {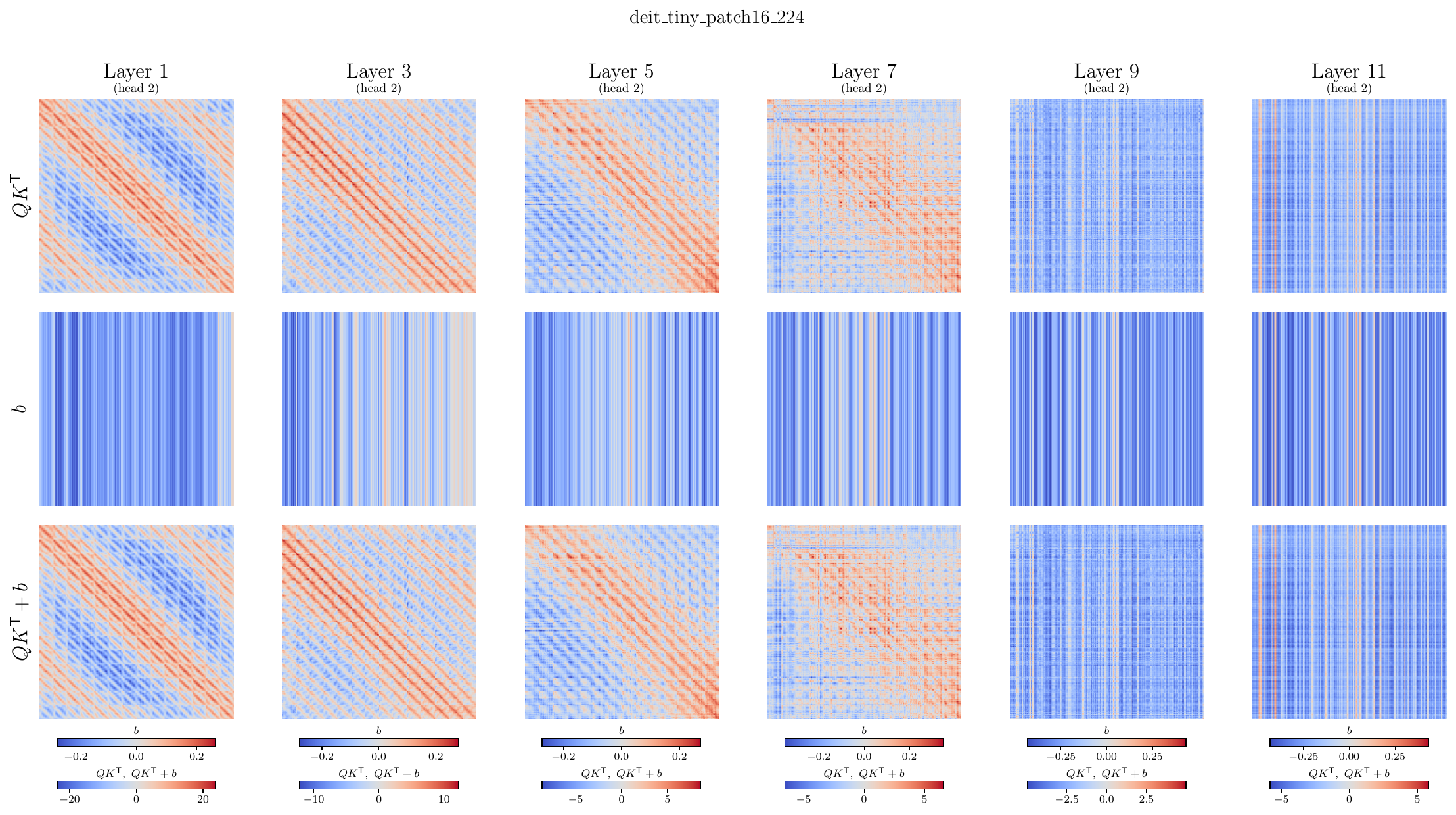}
    \caption{%
        Decomposition of the attention logits ($QK^\top$, $b$, $QK^\top+b$)
        for DeiT-Ti$_b$ at layers $\{1,3,5,7,9,11\}$, head~2.
        In the early layers, $QK^\top$ dominates with a strong diagonal
        pattern. With increasing depth the bias $b$ progressively captures the
        vertical structures, until in the final layers it accounts for most of
        the visible structure in the combined logits $QK^\top + b$.
    }
    \label{fig:decomp_tiny}
\end{figure*}

\begin{figure*}[p]
    \centering
    \includegraphics[width=\linewidth]%
        {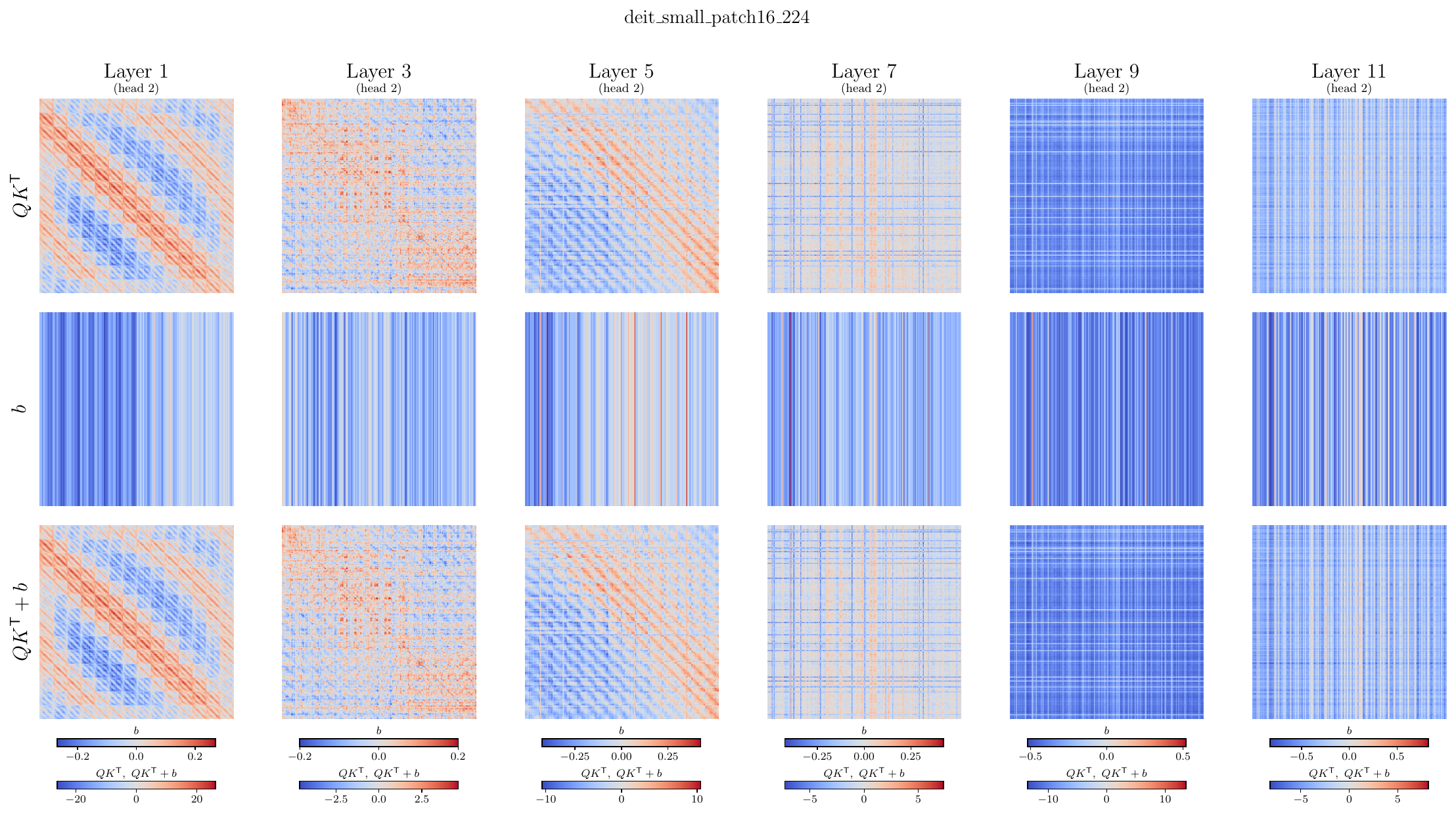}
    \caption{%
        Decomposition of the attention logits for DeiT-S$_b$ at layers
        $\{1,3,5,7,9,11\}$, head~2.
        The bias term increasingly dominates the combined logits with depth,
        consistently with \cref{fig:attention_decomposition} and \cref{fig:bias_qk_mean_scatter} of the main paper.
    }
    \label{fig:decomp_small}
\end{figure*}

\begin{figure*}[p]
    \centering
    \includegraphics[width=\linewidth]%
        {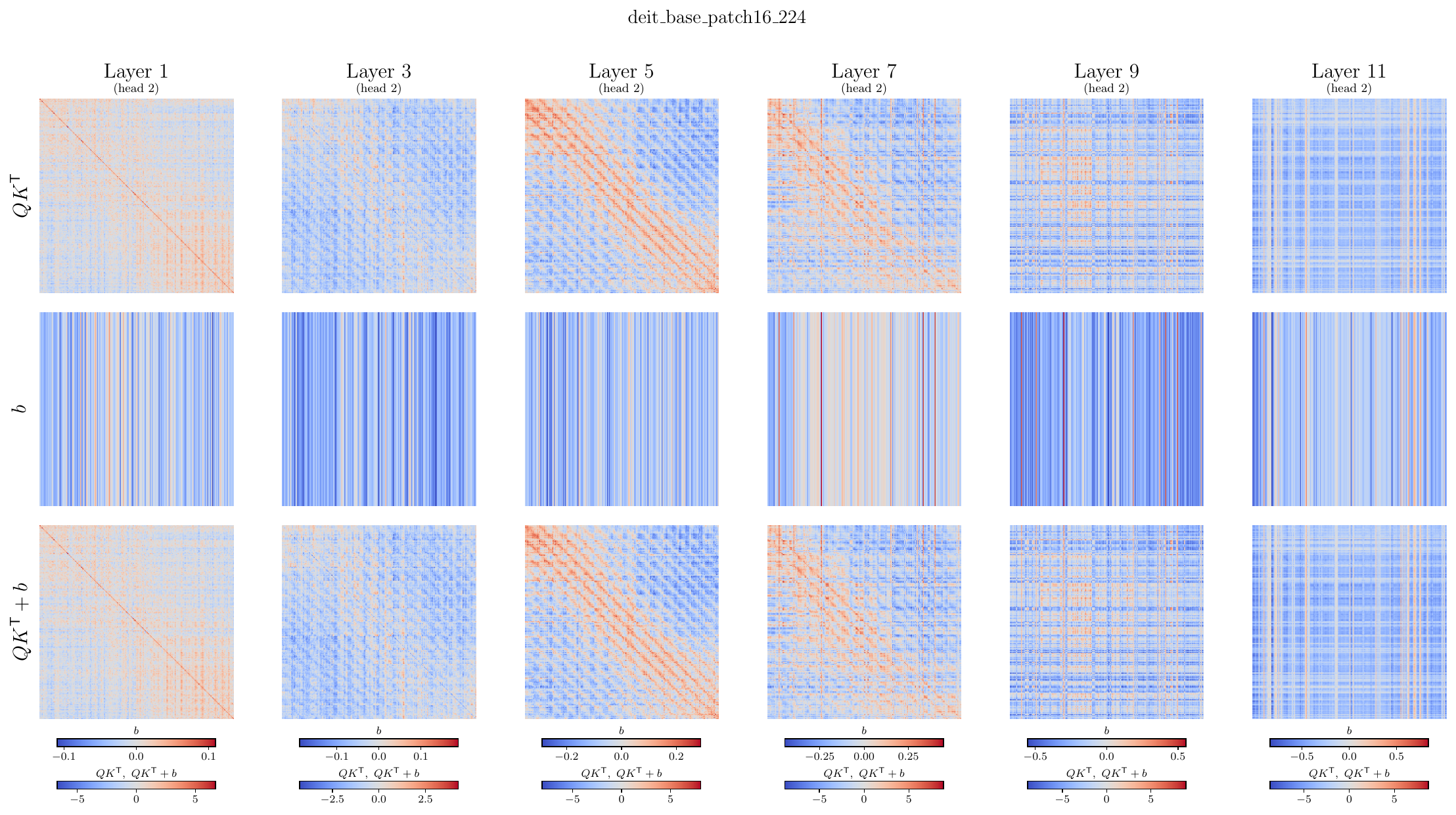}
    \caption{%
        Decomposition of the attention logits for DeiT-B$_b$ at layers
        $\{1,3,5,7,9,11\}$, head~2.
        The depth-wise takeover of the bias term is visible across all three
        model sizes, confirming that the learned bias captures a universal
        property of deep vision-transformer attention.
    }
    \label{fig:decomp_base}
\end{figure*}

\begin{figure*}[p]
    \centering
    \includegraphics[width=\linewidth]%
        {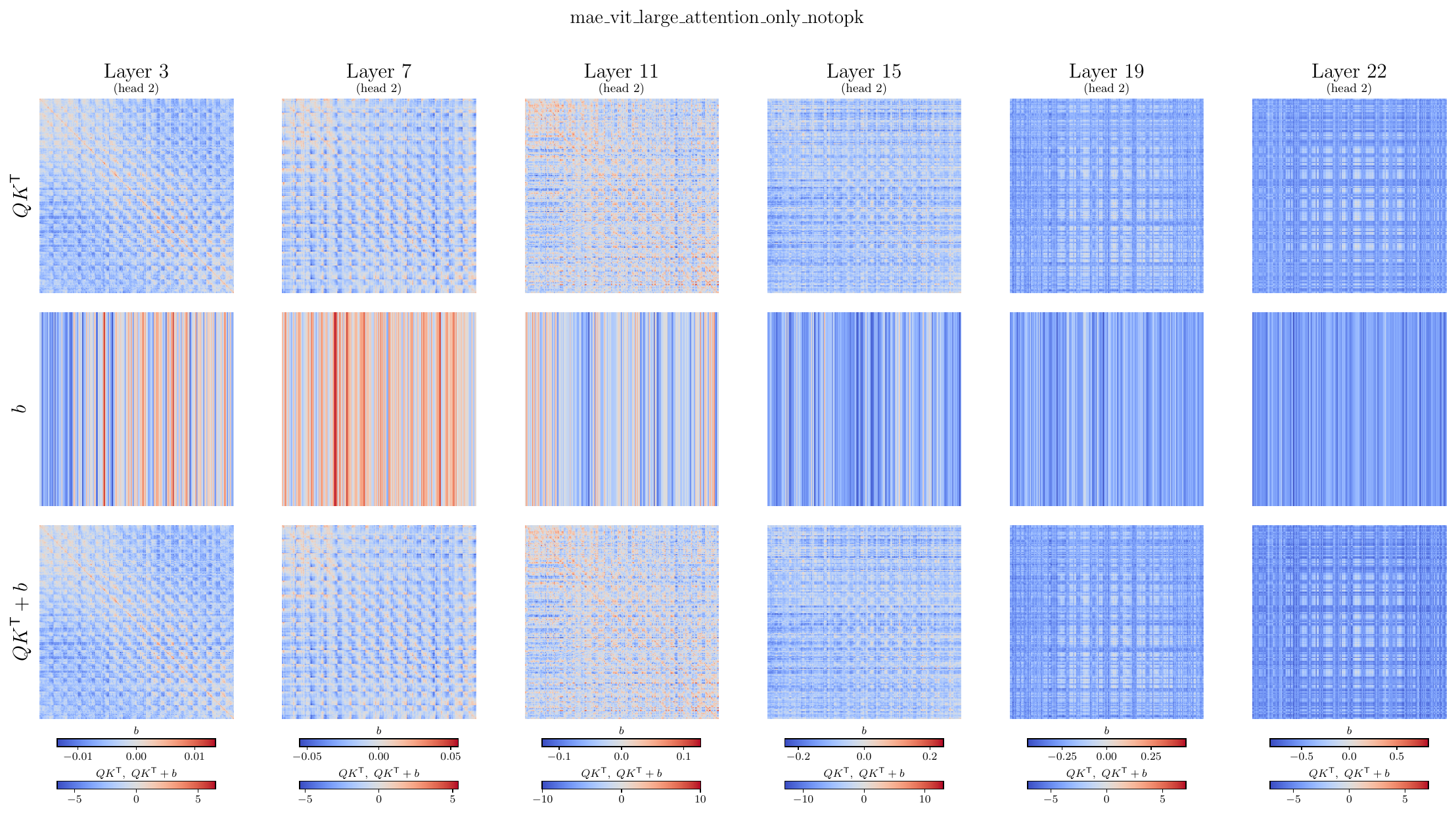}
    \caption{%
        Decomposition of the attention logits ($QK^\top$, $b$, and
        $QK^\top+b$) for MAE-pretrained ViT-L$_b$ at layers
        $\{3,7,11,15,19,22\}$, head~2. The model is fine-tuned with the bias
        term but without top-$k$ pruning. As depth increases, the bias
        progressively captures the vertical, query-independent structure of
        the combined logits, reproducing the behavior observed for DeiT.
    }
    \label{fig:decomp_vit_large}
\end{figure*}

\paragraph{Decomposition visualizations.}
\cref{fig:decomp_tiny,fig:decomp_small,fig:decomp_base} show the attention-logit
decomposition ($QK^\top$, $b$, $QK^\top + b$) for DeiT-Ti, DeiT-S, and DeiT-B
at layers $\{1, 3, 5, 7, 9, 11\}$ on a single validation image.
These extend \cref{fig:attention_decomposition} of the main paper to a denser set of depths and to multiple
architectures.
Across all models, the bias term $b$ progressively takes over the vertical
structure of the combined logits: in early layers $QK^\top$ dominates with its
diagonal pattern, while in later layers $b$ alone accounts for most of the
visible structure in $QK^\top + b$.
The transition is gradual and consistent with the growing depth-wise correlation
reported in \cref{fig:bias_qk_mean_scatter} of the main paper.
The decomposition of MAE-pretrained ViT-L$_b$ in \cref{fig:decomp_vit_large} exhibits the same depth-wise
behavior across its 24 blocks, extending this observation beyond the DeiT
family and to a different pretraining strategy.

\begin{figure*}[p]
    \centering
    \includegraphics[width=\linewidth]%
        {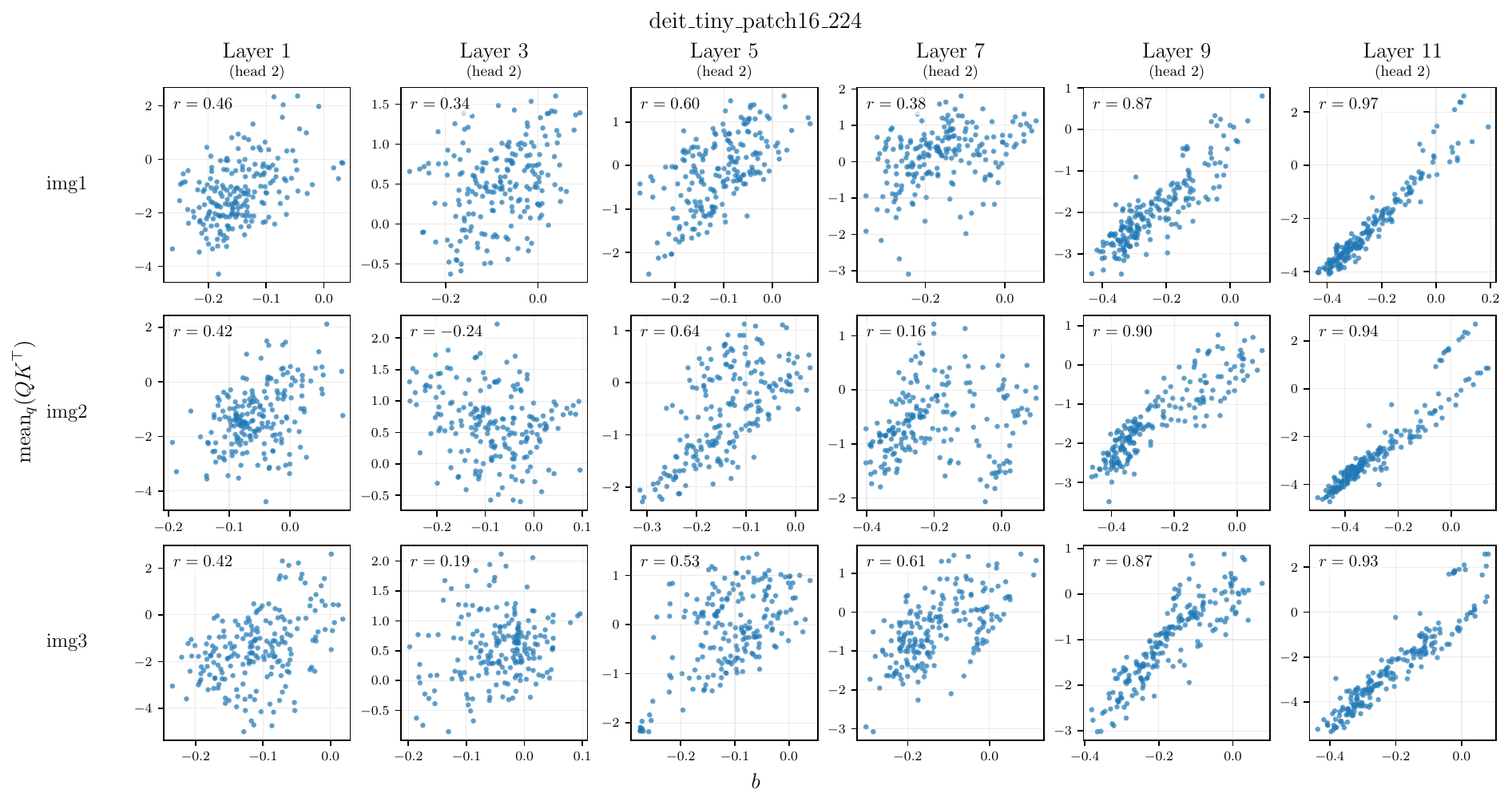}
    \caption{%
        Scatter plots of the learned bias $b$ versus
        $\operatorname{mean}_q(QK^\top)_k$ for DeiT-Ti$_b$ at layers
        $\{1,3,5,7,9,11\}$ (columns), head~2, across multiple validation
        images (rows).
        The Pearson correlation $r$ reported per panel grows steadily with
        depth and remains consistent across images, confirming that the bias
        provides a reliable surrogate across different input images of column
        importance.
    }
    \label{fig:scatter_tiny}
\end{figure*}

\begin{figure*}[p]
    \centering
    \includegraphics[width=\linewidth]%
        {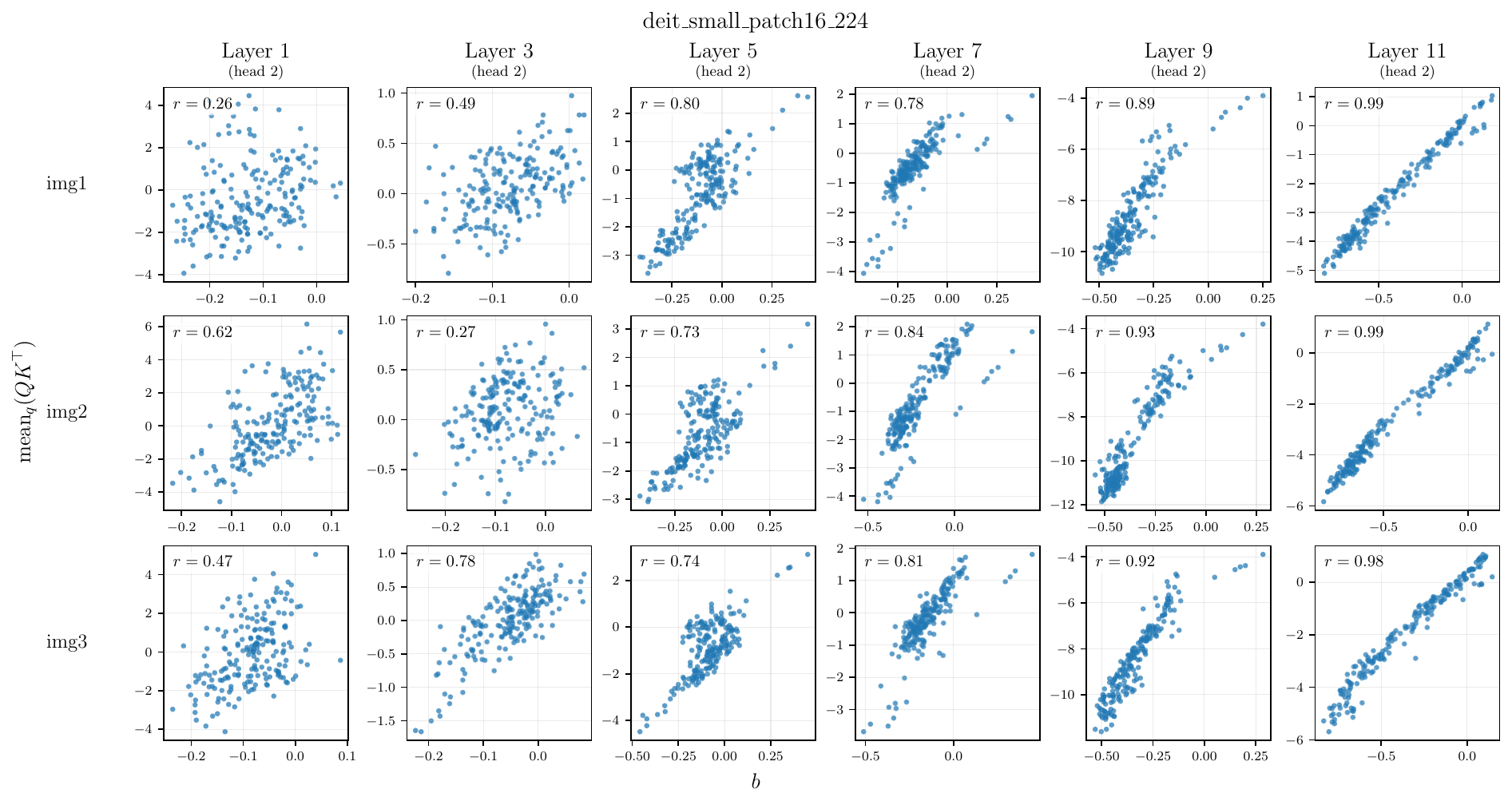}
    \caption{%
        Scatter plots for DeiT-S$_b$ at layers $\{1,3,5,7,9,11\}$, head~2,
        across multiple validation images.
        The correlation reaches $r \approx 0.98$ in the final layer,
        independently of the input image, showing that the bias approximates
        the query-averaged column importance with high fidelity in the deeper
        blocks.
    }
    \label{fig:scatter_small}
\end{figure*}

\begin{figure*}[p]
    \centering
    \includegraphics[width=\linewidth]%
        {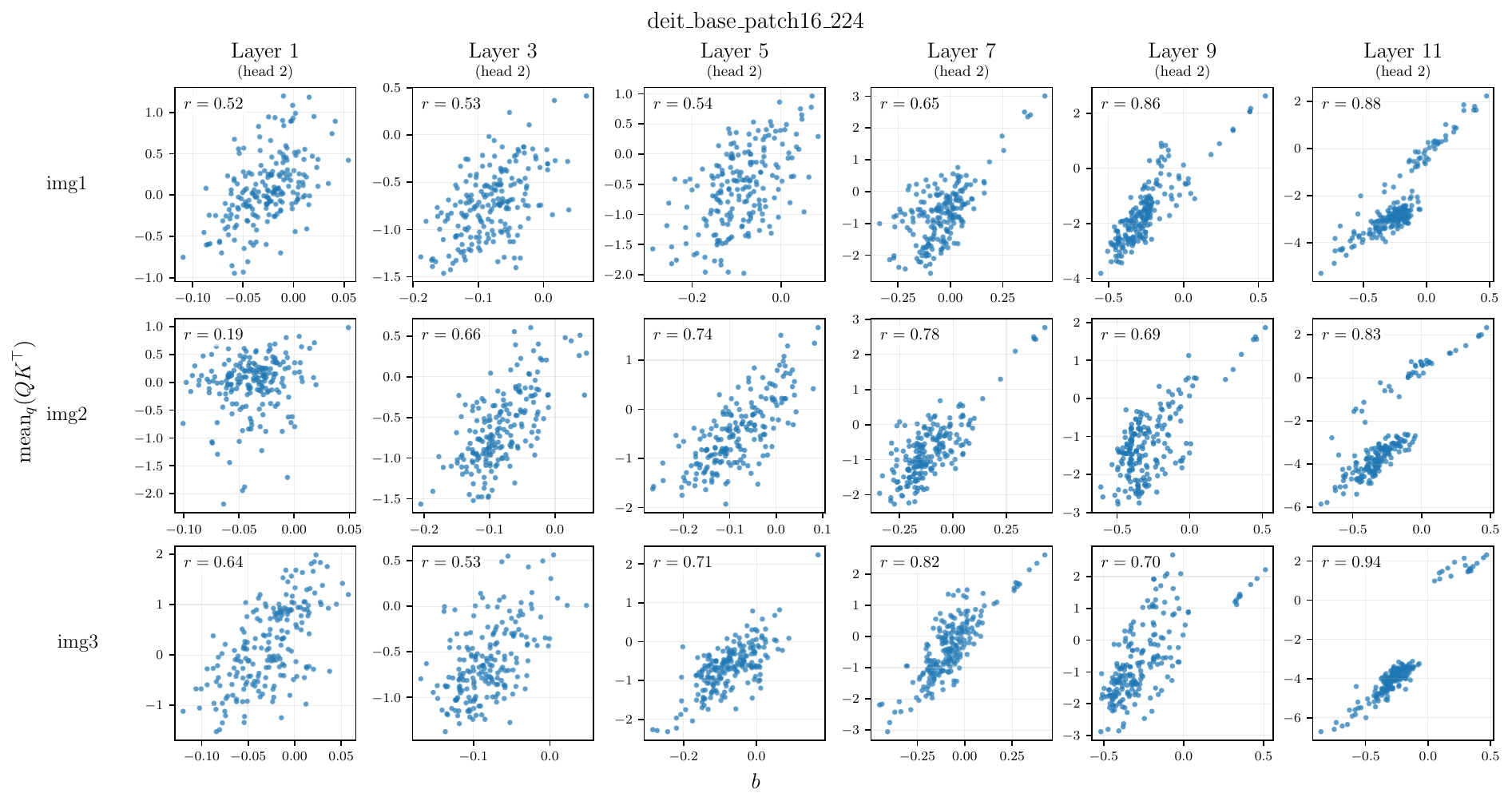}
    \caption{%
        Scatter plots for DeiT-B$_b$ at layers $\{1,3,5,7,9,11\}$, head~2,
        across multiple validation images.
        The depth-wise growth of $r$ is reproduced in the larger model,
        confirming that the relationship between the bias and the
        query-averaged column importance is a property of the attention
        mechanism rather than of a specific model scale.
    }
    \label{fig:scatter_base}
\end{figure*}

\begin{figure*}[p]
    \centering
    \includegraphics[width=\linewidth]%
        {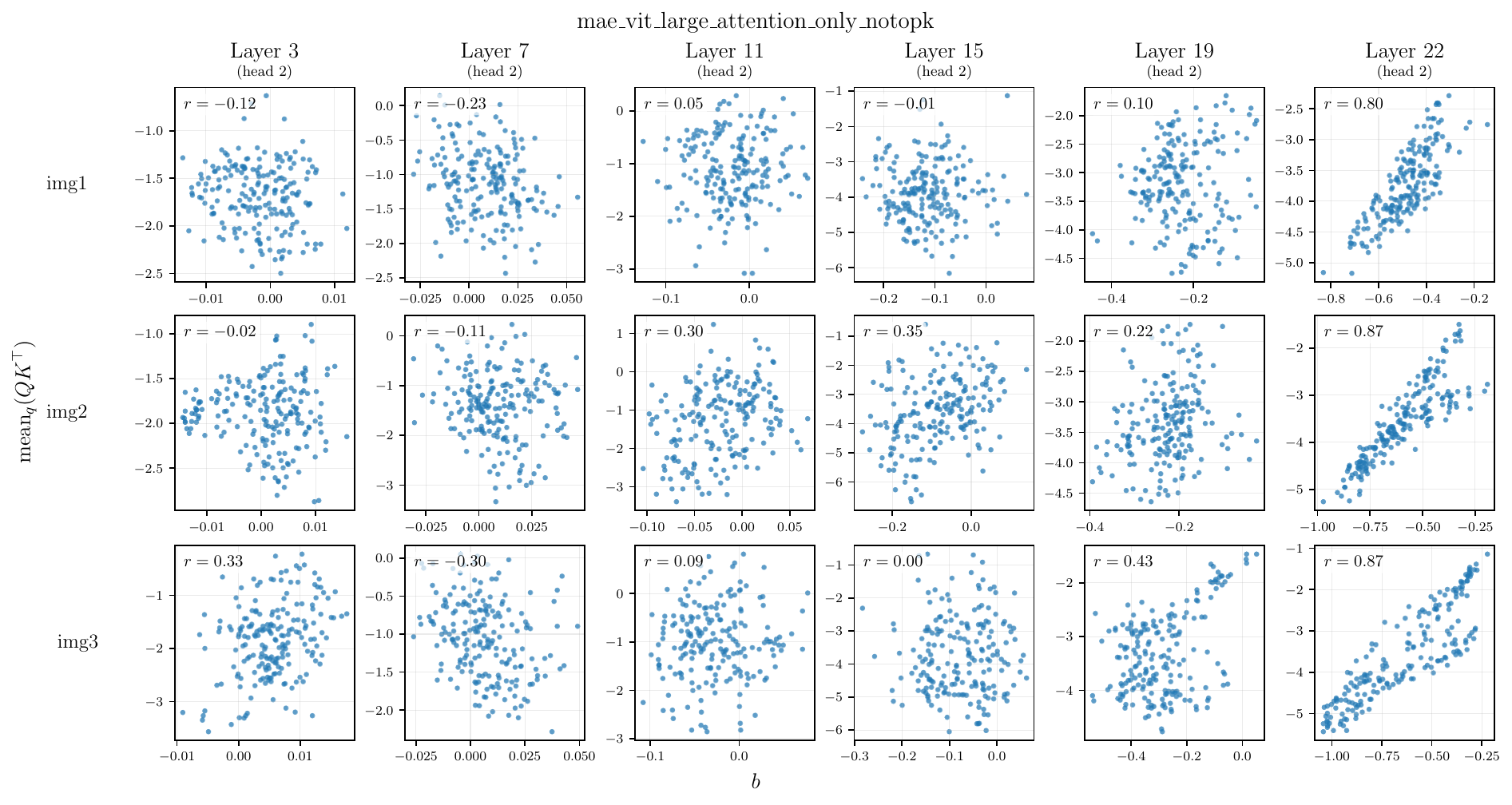}
    \caption{%
        Learned bias $b$ versus the per-column mean
        $\operatorname{mean}_q(QK^\top)_k$ for MAE-pretrained ViT-L$_b$ at
        layers $\{3,7,11,15,19,22\}$ (columns), head~2, across multiple
        ImageNet validation images (rows). The Pearson correlation reported in
        each panel becomes stronger with depth, showing that the bias also
        learns a reliable proxy for query-averaged column importance in ViT-L.
    }
    \label{fig:scatter_vit_large}
\end{figure*}

\paragraph{Scatter-plot visualizations.}
\cref{fig:scatter_tiny,fig:scatter_small,fig:scatter_base} extend the
per-column scatter analysis of \cref{fig:bias_qk_mean_scatter} to multiple validation images (rows) for
DeiT-Ti$_b$, DeiT-S$_b$, and DeiT-B$_b$.
Each panel plots the learned bias $b$ against the per-query mean of the
query-key logits, $\operatorname{mean}_q(QK^\top)$, and reports the Pearson
correlation $r$.
Across all models and all tested images, the correlation follows the same
depth-dependent trend: weak and variable in early layers (occasionally near zero or negative), increasing with depth to \(r \approx 0.83-0.99\) in the final blocks.
The consistency across images confirms that the bias learns a reliable
approximation of the column-wise query-averaged importance, independently of the
specific content of the input.
As shown in \cref{fig:scatter_vit_large}, we observe the same qualitative growth for ViT-L$_b$, showing
that the relationship between the learned bias and query-averaged column
importance also persists at larger scale.

\subsection{Correlation with pre-softmax column importance}

In \cref{subsec:prelim_experiments} of the main paper, we showed that the learned bias correlates with
the per-column mean of the pre-softmax query-key scores,
$\operatorname{mean}_q(QK^\top)$.
\cref{fig:pre-soft} extends this analysis by reporting the layer-wise Pearson
correlation computed over 200 ImageNet validation images and averaged over
attention heads considering our DeiT-S$_b$.

The correlation increases steadily with network depth, approaching unity in the
last transformer blocks.
This confirms that the learned bias progressively converges towards a cheap,
pre-computable approximation of the average column importance that would
otherwise require materializing the full $N \times N$ query-key matrix.

% ---- attention grid figures ----------------------------------------

% ---- decomposition figures -----------------------------------------

% ---- scatter-plot figures ------------------------------------------

%% file: sec/pseudo_code.tex
\begin{algorithm}[t]
\caption{Bias-based attention with column pruning (one head, one layer).}
\label{alg:pruned_attention}
\begin{algorithmic}[1]
\Require tokens $X \in \mathbb{R}^{N\times d}$; projections $W^Q,W^K,W^V,W^b$; selection rule $\mathcal{S}$
\State $b \gets X W^b$ \Comment{$b\in\mathbb{R}^N$: query-independent importance, computed first}
\State $\kept \gets \mathcal{S}(b)$ \label{algo:select} \Comment{indices of kept columns}
\State $Q \gets X W^Q$ \Comment{all queries are kept}
\State $K_{\kept} \gets X_{\kept} W^K$,\quad $V_{\kept} \gets X_{\kept} W^V$ \Comment{keys/values only for kept tokens}
\State $L \gets \dfrac{Q K_{\kept}^\top}{\sqrt{d_k}} + \onevec_N\, b_{\kept}^\top$ \Comment{logits over kept columns}
\State \Return $\softmax(L)\, V_{\kept}$
\end{algorithmic}
\end{algorithm}

%% file: tables/table_deit-tiny.tex
\begin{table}[t]
\centering
\scriptsize
\caption{Accuracy and efficiency of bias-based attention pruning on ImageNet-1k with different pruning schedules. Inference latency is measured with the math attention backend on a single $1024 \times 1024$ image on one NVIDIA A40 GPU.}
\label{tab:pruning_results_deit_tiny}
% \resizebox{\linewidth}{!}{%
% \begin{tabular}{lccccc}
\begin{tabular*}{\columnwidth}{
        @{\extracolsep{\fill}}
        lccccc
        @{}
    }
\toprule
\multirow{2}{*}{Model}
& \multirow{2}{*}{\shortstack{Pruning\\schedule}}
& \multicolumn{2}{c}{Accuracy}
& \multirow{2}{*}{\shortstack{FLOPs\\(G)}}
& \multirow{2}{*}{\shortstack{Latency\\(ms)}} \\
\cmidrule(lr){3-4}
&
& Top-1 (\%)
& Top-5 (\%)
&
% &
&
\\
\midrule

        DeiT-Ti
        & -
        & $\mathbf{72.13}$
        & $\mathbf{91.14}$
        & $199.4$
        & $47.92_{\pm 0.10}$ \\
        \midrule
        DeiT-Ti$_b$
        & \(0.9 \searrow 0.5\)
        & $71.47$
        & $90.55$
        & $153.0$
        & $42.22_{\pm 0.07}$ \\
        DeiT-Ti$_b$
        & \(0.9 \searrow 0.3\)
        & $70.32$
        & $90.11$
        & $137.6$
        & $37.29_{\pm 0.05}$ \\
        DeiT-Ti$_b$
        & \(0.9 \searrow 0.1\)
        & $69.10$
        & $89.73$
        & $\mathbf{122.1}$
        & $\mathbf{32.52_{\pm 0.13}}$ \\
        
        \bottomrule
    \end{tabular*}%
% }

\end{table}

%% file: tables/table_ablation_column.tex
\begin{table}[t]
\centering
\caption{Ablation of the column-selection strategy on ImageNet100 using ViT-S$_b$. The pruning ratio denotes the fraction of removed attention columns.}
\label{tab:selection_ablation}
\resizebox{\linewidth}{!}{%
\begin{tabular}{lcccc}
\toprule
\multirow{2}{*}{Method}
& \multirow{2}{*}{Selection rule}
& \multirow{2}{*}{\shortstack{Pruning\\ratio (\%)}}
& \multicolumn{2}{c}{Accuracy} \\
\cmidrule(lr){4-5}
&
&
& Top-1 (\%)
& Top-5 (\%) \\
\midrule

        No pruning
        & --
        & 0
        & 80.04
        & 93.30 \\
        \midrule
        Probability threshold
        & \(\tau=0.8\)
        & 49.55
        & 79.66
        & \textbf{93.18} \\
        Fixed top-\(k\)
        & \(0.5\)
        & 50.00
        & 79.70
        & 92.30 \\
        Layer-wise top-\(k\)
        & \(0.9\searrow0.1\)
        & 50.00
        & \textbf{80.48}
        & \textbf{93.18} \\
        
        \bottomrule
    \end{tabular}%
}
\end{table}

%% file: tables/table_ablation_prob.tex
\begin{table}[t]
\centering
\scriptsize
\caption{Results of the probability-threshold pruning strategy on ImageNet100. The pruning ratio denotes the fraction of removed attention columns and is measured at the end of training.}
\label{tab:score_threshold_imagenet100}
% \resizebox{\linewidth}{!}{%
% \begin{tabular}{lcccc}
\begin{tabular*}{\columnwidth}{
        @{\extracolsep{\fill}}
        lcccc
        @{}
    }
\toprule
\multirow{2}{*}{Model}
& \multirow{2}{*}{\shortstack{Probability\\threshold}}
& \multirow{2}{*}{\shortstack{Pruning\\ratio (\%)}}
& \multicolumn{2}{c}{Accuracy} \\
\cmidrule(lr){4-5}
&
&
& Top-1 (\%)
& Top-5 (\%) \\
\midrule

ViT-Ti
& --
& 0
& 79.08
& 92.90 \\
\midrule
ViT-Ti$_b$
& \cross
& 0
& 78.40
& 92.82 \\
ViT-Ti$_b$
& \(\tau=0.8\)
& 48.00
& 78.56
& 93.14 \\

\midrule
\midrule

ViT-S
& --
& 0
& 80.12
& 92.54 \\
\midrule
ViT-S$_b$
& \cross
& 0
& 80.04
& 93.30 \\
ViT-S$_b$
& \(\tau=0.8\)
& 49.55
& 79.66
& 93.18 \\
ViT-S$_b$
& \(\tau=0.7\)
& 61.58
& 78.50
& 92.40 \\
ViT-S$_b$
& \(\tau=0.6\)
& 70.09
& 78.76
& 91.92 \\

\bottomrule
\end{tabular*}%
% }
\end{table}

%% file: tables/table_ablation_topk.tex
\begin{table}[t]
\centering
\scriptsize
\caption{Results of the top-\(k\) pruning strategy on ImageNet100. The pruning ratio denotes the fraction of removed attention columns. The top-\(k\) value is expressed as the fraction of tokens that are kept.}
\label{tab:topk_imagenet100}
% \resizebox{\linewidth}{!}{%
% \begin{tabular}{lcccc}
\begin{tabular*}{\columnwidth}{
        @{\extracolsep{\fill}}
        lcccc
        @{}
    }
\toprule
\multirow{2}{*}{Model}
& \multirow{2}{*}{\shortstack{Top-\(k\)\\value}}
& \multirow{2}{*}{\shortstack{Pruning\\ratio (\%)}}
& \multicolumn{2}{c}{Accuracy} \\
\cmidrule(lr){4-5}
&
&
& Top-1 (\%)
& Top-5 (\%) \\
\midrule

ViT-Ti
& --
& 0
& 79.08
& 92.90 \\
\midrule
ViT-Ti$_b$
& \cross
& 0
& 78.40
& 92.82 \\
ViT-Ti$_b$
& \(0.9\searrow0.1\)
& 50
& 78.74
& 92.94 \\
ViT-Ti$_b$
& \(0.5\)
& 50
& 78.68
& 92.80 \\

\midrule
\midrule

ViT-S
& --
& 0
& 80.12
& 92.54 \\
\midrule
ViT-S$_b$
& \cross
& 0
& 80.04
& 93.30 \\
ViT-S$_b$
& \(0.9\searrow0.1\)
& 50
& 80.48
& 93.18 \\
ViT-S$_b$
& \(0.7\)
& 30
& 79.58
& 92.62 \\
ViT-S$_b$
& \(0.5\)
& 50
& 79.70
& 92.30 \\
ViT-S$_b$
& \(0.3\)
& 70
& 78.44
& 92.42 \\

\bottomrule
\end{tabular*}%
% }
\end{table}

%% file: tables/table_bias_only.tex
\begin{table}[t]
\centering
\scriptsize
\caption{ImageNet100 accuracy when replacing the query–key term with the broadcast bias in the last transformer blocks. ``Query-key-free blocks'' denote the number of final blocks using only \(\mathbf{1}b^\top\).}
\label{tab:bias_only_imagenet100}
% \resizebox{\linewidth}{!}{%
\begin{tabular*}{\columnwidth}{
    @{\extracolsep{\fill}}
    lccc
    @{}
}
\toprule
\multirow{2}{*}{Model}
& \multirow{2}{*}{\shortstack{Query-key-free\\blocks}}
& \multicolumn{2}{c}{Accuracy} \\
\cmidrule(lr){3-4}
&
& Top-1 (\%)
& Top-5 (\%) \\
\midrule

        ViT-S
        & --
        & $80.12$
        & $92.54$ \\
        \midrule
        ViT-S$_b$
        & \cross
        & $80.04$ %{\small\((-0.08)\)}
        & $\mathbf{93.30}$ \\%{\small\((+0.76)\)} \\
        ViT-S$_b$
        & 4
        & $\mathbf{80.28}$ %{\small\((+0.16)\)}
        & $92.82$ \\%{\small\((+0.28)\)} \\
        ViT-S$_b$
        & 6
        & $79.30$ %{\small\((-0.82)\)}
        & $92.72$ \\%{\small\((+0.18)\)} \\
        ViT-S$_b$
        & 8
        & $79.22$ %{\small\((-0.90)\)}
        & $92.08$ \\%{\small\((-0.46)\)} \\
        \bottomrule
    \end{tabular*}%
% }

\end{table}